\documentclass{article}

\newif\ifreview 
\newif\ifarxiv 
\newif\ifcamera 
\newif\ifrebuttal

\newcommand{\X}{\boldsymbol{X}}

\newcommand{\x}{\boldsymbol{x}}

\newcommand{\T}{{\!\top\!}}

\newcommand{\tX}{\underline{\bm X}}

\usepackage{calc,amssymb,bm,url,color,theorem,epstopdf,nicefrac, booktabs}
\usepackage[numbers]{natbib}

\newtheorem{Prop}{Proposition}

 \usepackage[preprint]{neurips_2025}

\usepackage[utf8]{inputenc} 
\usepackage[T1]{fontenc}    
\usepackage{hyperref}       
\usepackage{url}            
\usepackage{booktabs}       
\usepackage{amsfonts}       
\usepackage{nicefrac}       
\usepackage{microtype}      
\usepackage{xcolor}         
\usepackage{graphicx,enumitem,pifont}
\usepackage{amsmath} 
\usepackage{subcaption}

\title{Pseudo-label Induced Subspace Representation Learning for Robust Out-of-Distribution Detection}

\author{%
  Tarhib Al Azad, Faizul Rakib Sayem and Shahana Ibrahim\\
  Department of Electrical and Computer Engineering\\
  University of Central Florida\\
  Orlando, FL 32826 \\
}

\begin{document}

\maketitle

\begin{abstract}

Out-of-distribution (OOD) detection lies at the heart of robust artificial intelligence (AI), aiming to identify samples from novel distributions beyond the training set. Recent approaches have exploited feature representations as distinguishing signatures for OOD detection. However, most existing methods rely on restrictive assumptions on the feature space that limit the separability between in-distribution (ID) and OOD samples. In this work, we propose a novel OOD detection framework based on a pseudo-label-induced subspace representation, that works under more relaxed and natural assumptions compared to existing feature-based techniques. In addition, we introduce a simple yet effective learning criterion that integrates a cross-entropy-based ID classification loss with a subspace distance-based regularization loss to enhance ID-OOD separability. Extensive experiments validate the effectiveness of our framework.
\end{abstract}    
\section{Introduction}
\label{sec:intro}

Deep learning models have exhibited remarkable performance across various domains, ranging from computer vision to natural language processing. However, a key challenge emerges when deploying these models in real-world scenarios: they struggle to detect data samples from unknown distributions, often producing overconfident and potentially unreliable predictions \cite{goodfellow2014explaining}. Therefore, addressing the task of out-of-distribution (OOD) detection is crucial for reliable artificial intelligence (AI), especially in safety critical applications such as autonomous driving \cite{geiger2012we} and medical diagnosis \cite{thomas2017unsupervised}. The goal of OOD detection is not only to ensure accurate predictions for data from known distributions but also to identify instances from unseen distributions that are not encountered during training \cite{hendrycks2016baseline}.

OOD detection remains an active research area within AI over the last decade---see a recent survey in \cite{yang2024generalized}. The detection of semantic shift (i.e., the occurrence of new classes) is central to the OOD detection tasks, where the label space is different between in-distribution (ID) data (i.e., the training data) and OOD data. Early approaches focused on utilizing the output of the trained neural networks to derive detection scores to identify the OOD samples, e.g., \cite{hendrycks2016baseline, liang2018enhancing} used the softmax outputs from the networks to characterize the ``OOD-ness" of the data samples. Nonetheless, deep neural networks tend to be overconfident on OOD samples, that often leads to poor ID-OOD separability in these methods.  This motivated several approaches to utilize the pre-softmax activation weights of the neural networks for detection \cite{hendrycks2022scaling,sun2022dice,sun2021react,dong2022neural}. Yet, they still suffer from overconfidence issues and are sensitive to the neural network architectures.

Recently, distance-based methods have been gaining popularity for OOD detection tasks \cite{lee2018simple,sun2022deepnn,ming2023hypersphericalood,SehwagSSD,ghosal2023overcomeood}. 
These approaches operate on the assumption that the feature representations extracted from OOD data tend to lie farther from the ID feature space. Since deep neural networks inherently encode semantic similarity in their feature embeddings, i.e., forming clusters for similar samples, distance-based methods can define structured feature spaces for ID data to distinguish OOD samples more effectively.
For instance, the work in \cite{lee2018simple} proposed modeling the feature embedding space of ID data as a mixture of multivariate Gaussian distributions and employs the Mahalanobis distance \cite{mahalanobis2018generalized} to quantify a sample’s deviation from the nearest class centroid for OOD detection. More recently, non-parametric metrics based on $k$-\textit{nearest neighbors} ($k$NN) distances have emerged as prominent OOD detection scores due to their distribution-free nature and improved performance on complex datasets \cite{sun2022deepnn}.
Several works have adopted this scoring scheme, focusing on improving the quality of the feature embedding space—particularly its clusterability—to enhance the detection \cite{ghosal2023overcomeood,ming2023hypersphericalood,SehwagSSD}. For example, \cite{SehwagSSD} employed supervised contrastive loss \cite{khosla2020supervised} to learn more discriminative feature representations.
Similarly, the work in \cite{ming2023hypersphericalood} encouraged intra-class compactness and inter-class separation within a \textit{hyperspherical} feature space to enhance the $k$NN-based detection, although by imposing  a Gaussian-like distribution assumption on the feature distribution. Another recent attempt \cite{ghosal2023overcomeood} proposed a mini-max optimization framework to extract a more representative low-dimensional subspace, focusing on the most discriminative features. Nonetheless, the search space for such a subspace is infinitely large, which limits the method’s ability to consistently identify an optimal representation for OOD detection.

\begin{figure}[t]
  \centering
  \begin{subfigure}[b]{0.48\linewidth}
    \centering
    \includegraphics[width=1.1\linewidth]{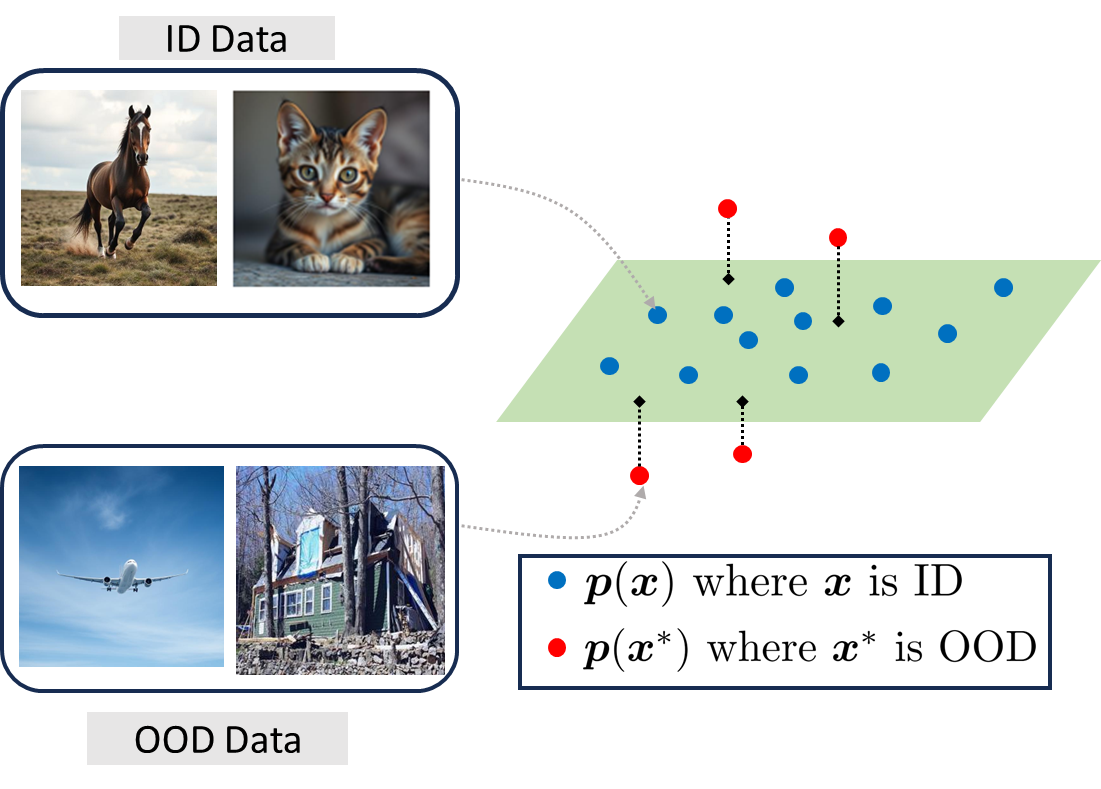}
    \caption{}
    \label{fig:ood:a}
  \end{subfigure}\hfill
  \begin{subfigure}[b]{0.48\linewidth}
    \centering
    \includegraphics[width=0.7\linewidth]{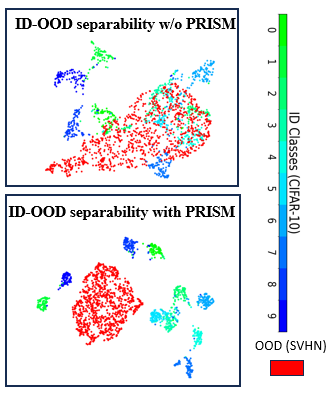}
    \caption{}
    \label{fig:ood:b}
  \end{subfigure}\hfill

   \caption{(a) The proposed OOD detection idea. Here, the subspace (shaded in green) is induced by the pseudo-label vector $\bm p(\bm x)$, which is a concatenation of different pseudo label probability vectors learned by the deep neural network.  The vectors $\bm p(\bm x)$ corresponding to the ID data lie on the subspace, whereas for OOD data, they are more likely to fall outside. (b) the UMAP visualizations \cite{mcinnes2018umap} of the penultimate feature vectors learned using the proposed {\bf \underline{P}}seudo-label {\bf \underline{R}}epresentation {\bf \underline{I}}nduced {\bf \underline{S}}ubspace {\bf \underline{M}}odeling (PRISM) approach, showing improved ID-OOD clusterability. Here, the top figure is obtained by learning the feature vectors directly from the labeled ID data without leveraging the pseudo-label subspace-based representation.   }
  \label{fig:ood_comparison}
\end{figure}

\noindent
\textbf{Our Contributions.} In this work, we propose a novel OOD detection framework, that leverages the subspace induced by a set of pseudo-labels---that are generated from extracted features during training. Our key finding reveals that the posterior probability distribution vectors of these pseudo-labels reside in a subspace spanned by their confusion matrices with respect to the ground-truth labels. Unlike recent approaches \cite{ming2023hypersphericalood,ghosal2023overcomeood}, which enforce feature representations to occupy unrelated subspaces, our method naturally derives the subspace from the pseudo-labels without imposing any distributional assumptions. Consequently, the learned feature space exhibits superior clusterability for OOD detection, enhancing the nonparametric detection schemes such those based on $k$NN---see Fig. \ref{fig:ood_comparison}. Our key contributions are as follows:
\vspace{-0.3em}
\begin{itemize}[left=0pt,label=\ding{228}]
    \item \textit{Novel OOD detection framework}: We propose a novel distance-based OOD detection framework leveraging a low-dimensional subspace induced by pseudo-labels, which are extracted from feature embeddings during training and are learned without incurring much computational overhead.
    \item \textit{End-to-end, regularized learning criterion}: We design an easy-to-handle, end-to-end learning criterion that jointly trains a label predictor for ID classification while learning discriminative features for effective OOD detection at test time. We also introduce a subspace distance-based regularization loss that promotes ID feature representations to reside in the designed low-dimensional subspace, thus further enhancing the ID-OOD separability.
    \item \textit{Promising empirical evidence for OOD detection}: We conduct extensive experiments using real-world datasets, e.g., on CIFAR-10 and CIFAR-100, and evaluate our approach across diverse OOD datasets. Additionally, we perform detailed ablation study on key hyperparameters to demonstrate the robustness of our approach.
\end{itemize}

\noindent
{\bf Notation.} Notations are defined in the supplementary materials in Sec. \ref{app:notation}.

\section{Problem Statement}
\label{sec:background}


Consider the input feature space ${\mathcal X} \subset \mathbb{R}^D$, where $D$ is the feature dimension. Let the label space ${\mathcal Y}=\{1,\dots,K\}$ denote the possible classes for the ID data. We consider the training set ${\mathcal D}$ such that
 \begin{align*}
     {\cal D} &= \{(\bm x_n,y_n)\}_{n=1}^N, ~\bm x_n \in {\cal X}, ~y_n \in {\cal Y},
 \end{align*}
 where $\bm x_n$ is the input feature vector corresponding to $n$th sample,  $y_n$ denotes the corresponding \textit{ground-truth} label, and $N$ is the total number of samples in the training set.
 Each element $(\bm x_n,y_n)$ is independent and identically distributed (i.i.d.) and is drawn uniformly at random from a joint data distribution ${\cal P}_{{\cal X}{\cal Y}}$. Let $\bm h: \mathbb{R}^D \rightarrow \mathbb{R}^L$ denote a deep neural network (DNN) trained on the training samples such that $\bm h(\bm x_n)$ is the $L$-dimensional encoded representation of the input feature vector $\bm x_n$. In multi-class classification setting, we also train a classifier head $\bm c: \mathbb{R}^{L} \rightarrow  \Delta^{K}$ in order to obtain the  label predictions, i.e., the label predictor is given by
     $\bm f(\bm x_n) = \bm c(\bm h(\bm x_n))$. 

\noindent
{\bf OOD Detection.} Traditionally, AI models are trained under the closed-world assumption that, they will encounter only the samples drawn from the same distribution as the training data during testing. However, in real-world scenarios, models are often exposed to samples that do not belong to the training distribution, referred as the OOD samples \cite{hendrycks2016baseline}. In classification settings,  there may occur a semantic-shift such that test samples may belong to an \textit{unknown} label space $ {\cal Y}^o$, where ${\cal Y} \cap {\cal Y}^o = \emptyset$. The objective of OOD detection is to determine whether a given test sample belongs to the ID distribution or not, thereby preventing models from assigning high-confidence predictions to OOD samples. In essence, OOD detection is a binary classification task aimed at distinguishing OOD samples from ID samples. Formally, this can be expressed using a detection function:
 $\bm g_{\tau}: \mathbb{R}^D \rightarrow \{{\sf ID},{\sf OOD}\}$ such that
 \begin{align} \label{eq:detect}
     \bm g_{\tau}(\bm x) = \begin{cases} {\sf ID} & s(\bm x) \ge \tau \\
     {\sf OOD} & s(\bm x) < \tau,
     \end{cases}
 \end{align}
 where $s(\bm x)$ is a scoring function associated with the input feature $\bm x$ and $\tau$ is the threshold. 
 Hence, the goal of OOD detection is two-fold: \textit{(i)} correctly predict the classes of the {ID} data samples (i.e., a well generalized $\bm f$) \textit{(ii)} correctly detect {OOD} data samples (i.e., accurate predictor $\bm g_{\tau}$).

\section{Proposed Approach}
Our goal is to design a feature representation space that can well-differentiate OOD samples from ID samples. Unlike the most distance-based detection strategies that rely on strong distributional assumptions on the representation space, we design a feature space that is more natural to the underlying classification task.  

\subsection{Pseudo-label Induced Subspace } \label{sec:plabel}
Consider the feature embedding $\bm h(\bm x_n) \in \mathbb{R}^L$ for the $n$th input sample (often obtained by the feature vectors from the penultimate layer of the deep neural network encoder). We first generate $M$ pseudo labels from this feature vector. Towards this, consider a projection head $\bm z: \mathbb{R}^{L} \rightarrow \mathbb{R}^{MK}$ such that we obtain the $MK$-dimensional feature representation as follows:
$$\tilde{\bm h}(\bm x_n) = \bm z(\bm h(\bm x_n)) = [\tilde{\bm h}_1(\bm x_n)^{\top}, \dots, \tilde{\bm h}_M(\bm x_n)^{\top}]^{\top},$$ where each $\tilde{\bm h}_m(\bm x_n) \in \mathbb{R}^K$ is the $m$th block vector in $\tilde{\bm h}(\bm x_n)$. To generate $M$ pseudo label probability vectors, we project each vector 
 $\tilde{\bm h}_m(\bm x_n)$ to the probability simplex $\Delta_K$ (e.g., by using a softmax operation), i.e.,
 \begin{align*}
     \bm p_m(\bm x_n) = \bm \sigma(\tilde{\bm h}_m(\bm x_n)),
 \end{align*}
where $\bm \sigma(\cdot)$ denotes the softmax operator.
 Note that each $\bm p_m(\bm x_n)$ represents a distinct pseudo label vector as it is derived from different sets of feature values in $\tilde{\bm h}(\bm x_n)$.
Intuitively, each $\bm p_m(\bm x_n)$ can be interpreted as the label distribution associated with $m$th pseudo label, denoted as $\widehat{y}_n^{(m)}$, i.e.,
\begin{align} \label{eq:pseudo_label}
    [\bm p_m(\bm x_n)]_{k} = {\sf Pr}(\widehat{y}_n^{(m)}=k|\bm x_n), \forall k \in [K].
\end{align}
We further have the following relation for this probability distribution:
\begin{align} 
    {\sf Pr}(\widehat{y}_n^{(m)}=k'|\bm x_n) &= 
    \sum_{k=1}^K {\sf Pr}(\widehat{y}_n^{(m)}=k'|y_n=k,\bm x_n)   {\sf Pr}(y_n=k|\bm x_n), \forall k' \in [K].\label{eq:bayes_rule}
\end{align}
where we have used the law of total probability and the Bayes' rule to derive the right-hand side.
The conditional probability term ${\sf Pr}(\widehat{y}_n^{(m)}=k'|y_n=k,\bm x_n)$ in \eqref{eq:bayes_rule} represents the confusions (or deviations) in pseudo label distribution relative to the ground-truth labels and the input feature (instance) vector.  
We show the instance-independence nature of this term using the following set of relations:
\begin{align}
    {\sf Pr}(\widehat{y}_n^{(m)}=k'|y_n=k,\bm x_n) 
    &= \frac{ {\sf Pr}(\widehat{y}_n^{(m)}=k',\bm x_n|y_n=k)}
    { {\sf Pr}(\bm x_n|y_n=k)} \notag \\
    &= \frac{ {\sf Pr}(\widehat{y}_n^{(m)}=k'|y_n=k)}
    { {\sf Pr}(\bm x_n|y_n=k)} {\sf Pr}(\bm x_n|\widehat{y}_n^{(m)}=k',y_n=k) \notag \\
    &= {\sf Pr}(\widehat{y}_n^{(m)}=k'|y_n=k), \label{eq:py}
\end{align}
where the first and second equalities are obtained by applying the Bayes' rule and the last equality is obtained by assuming
\begin{align} \label{eq:assump}
    {\sf Pr}(\bm x_n|\widehat{y}_n^{(m)}=k',y_n=k)={\sf Pr}(\bm x_n|y_n=k).
\end{align}
Note that the instance-independence is a widely used assumption in noisy label learning literature \cite{li2021provably,tanno2019learning,ibrahim2023deep}. Although the assumption in \eqref{eq:assump} is debatable,  it is intuitive in our setting as given the ground-truth label $y_n$, the pseudo label information $\widehat{y}_n^{(m)}=k'$ adds very little information about the distribution of the input features.
Consequently, applying the result \eqref{eq:py} in \eqref{eq:bayes_rule}, we get the following relation:
\begin{align} \label{eq:prob}
    {\sf Pr}(\widehat{y}_n^{(m)}=k'|\bm x_n)  =\sum_{k=1}^K {\sf Pr}(\widehat{y}_n^{(m)}=k'|y_n=k){\sf Pr}(y_n=k|\bm x_n).
\end{align}
One can define a pseudo-label \textit{confusion matrix} for each $m$ as
$$[\bm A_m]_{k',k} \triangleq {\sf Pr}(\widehat{y}_n^{(m)}=k'|y_n=k), \forall m \in [M], \forall k,k' \in [K] $$
that encapsulates all confusion probabilities into a $K \times K$-sized matrix $\bm A_m$. These confusion matrix terms---that frequently appear in noisy label learning frameworks \cite{dawid1979maximum,tanno2019learning,ibrahim2023deep}---act as correction terms for the noisy pseudo labels while learning the ground-truth label classifier. Also, note that they satisfy the probability simplex constraints on its columns, i.e.,
$\bm 1^{\top} \bm A_m = \bm 1^{\top}, \bm A_m \ge \bm 0.$
Defining the ground-truth classifier function $\bm f: \mathbb{R}^D \rightarrow \mathbb{R}^K$ such that $[\bm f(\bm x_n)]_k = {\sf Pr}(y_n=k|\bm x_n), \forall k \in [K] $, we derive the following generative model from \eqref{eq:prob}:
\begin{align} \label{eq:pm}
    \bm p_m(\bm x_n) = \bm A_m \bm f(\bm x_n), \forall m.
\end{align}
By stacking the probability vectors $\bm p_m(\bm x_n)$ for all $m$, we further have a factorization form as follows:
\begin{align} 
    \underbrace{\begin{bmatrix}\bm p_1(\bm x_n) \\ \vdots \\ \bm p_M(\bm x_n)\end{bmatrix}}_{\bm p(\bm x_n)} & =  \begin{bmatrix}\bm A_1 \bm f(\bm x_n) \\ \vdots \\ \bm A_M \bm f(\bm x_n)\end{bmatrix} = \underbrace{\begin{bmatrix}\bm A_1  \\ \vdots \\ \bm A_M \end{bmatrix}}_{\bm W}\bm f(\bm x_n). \label{eq:subspace}
\end{align}
The relation in \eqref{eq:subspace} implies that the pseudo label representation $\bm p(\bm x_n) \in \mathbb{R}^{MK}$ lies in a low-dimensional subspace spanned by the columns of the confusion stacked matrix $\bm W \in \mathbb{R}^{MK \times K}$, denoted as ${\cal R}(\bm W)$. This implies that even if the feature vectors $\bm p(\bm x_n)$'s are high dimensional, there exists a low-dimensional subspace that they reside, provided $M>1$. Such a nice geometry promotes a better ID-OOD separability in the feature space, as we will further see in our experiments.

\begin{figure*}[t]
    \centering
    \includegraphics[width=0.99\linewidth]{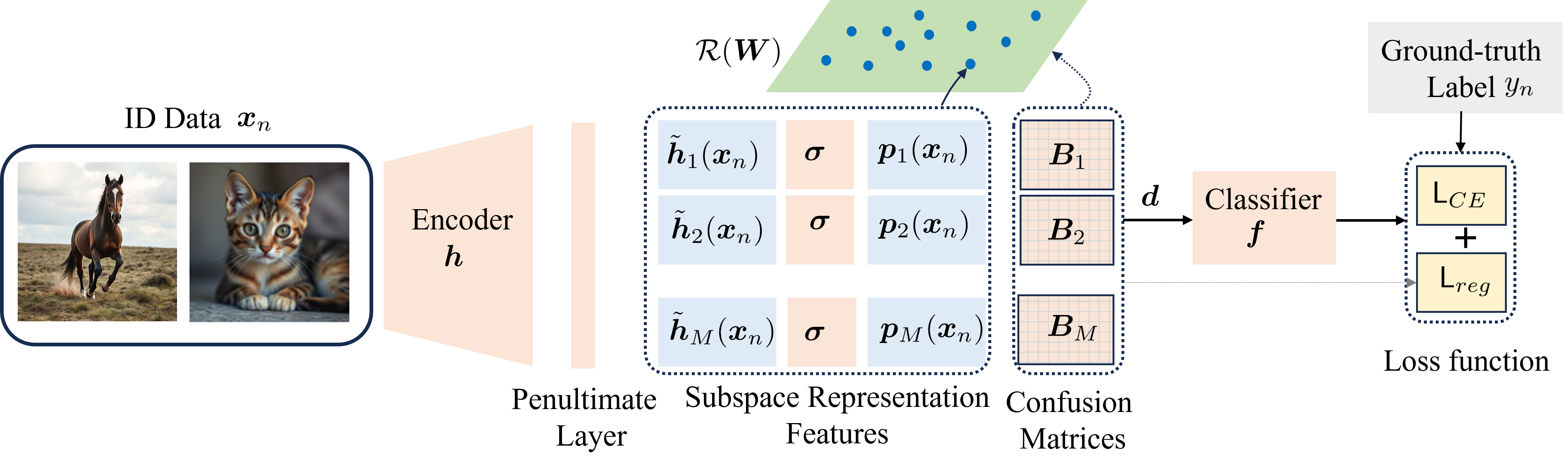}
    \caption{The proposed PRISM framework, and the training pipeline using the cross entropy loss and the subspace distance-based regularization loss. The green shaded region represents the subspace spanned by the columns of the matrix $\bm W= [\bm B^{-\top}_1,  \bm B^{-\top}_2, \dots, \bm B^{-\top}_M]^{\top}$, where $\bm B_m$'s are confusion matrices associated with the pseudo labels.} 
    \label{fig:PRISM}
    
\end{figure*}

\subsection{Learning Criterion}
In this section, we outline our proposed training strategy that promotes learning of the subspace representations as derived in the previous section. 
Towards this, we design an end-to-end learning criterion that learns the feature vectors $\bm p(\bm x_n)$ to help promote better ID-OOD separability. Our proposed loss function design involves two key objectives: \textit{(i)} the output predictions $\bm f(\bm x_n)$'s for each ID sample are correctly assigned to the corresponding ground-truth label \textit{(ii)} the pseudo label vectors $\bm p(\bm x_n)$'s for each ID sample lie on the subspace spanned by the columns of $\bm W$, i.e., ${\cal R}(\bm W)$.

\noindent
{\bf Constrained cross-entropy loss.}
To accomplish the first objective \textit{(i)}, we consider the following constrained learning criterion:
\begin{subequations} \label{eq:criterion_ce}
\begin{align}
    \underset{\{\bm B_m\}, \bm d, \bm \theta}{\rm minimize}&~ -\frac{1}{N}\sum_{n=1}^N \sum_{k=1}^K\mathbb{I}[y_n=k]\log ([\bm B\bm p_{\bm \theta}(\bm x_n)]_k) \label{eq:objective}\\
    {\rm subject~to}&~\bm B = [d_1\bm B_1, d_2 \bm B_2, \dots, d_M\bm B_M],\label{eq:const1}\\
    &~\bm d =[d_1,\dots, d_M]^{\top} \ge \bm 0,~ \bm 1^{\top}\bm d=1,\label{eq:const2}\\
    &~\bm 1^{\top} \bm B_m = \bm 1^{\top}, \bm B_m \ge \bm 0, \forall m, \label{eq:const3}
\end{align}
\end{subequations}
where $\bm B_m \in \mathbb{R}^{K \times K}$, $\bm d \in \mathbb{R}^M$, and $\bm \theta$ denotes the neural network parameters including the encoder and the projection head. Here, we employed a reparameterization strategy by introducing the $\bm d$ vector whose entries act as weight coefficients for the $\bm B_m$ matrices. Consequently, the constraint \eqref{eq:const1} ensures that the term $\bm B\bm p_{\bm \theta}(\bm x_n)$ in \eqref{eq:objective} represents a probability vector of size $K$. The constraints in \eqref{eq:const1}-\eqref{eq:const2} enforce the probability simplex constraints on $\bm B_m$'s and $\bm d$, respectively.
To establish the feasibility of the optimization problem in \eqref{eq:criterion_ce}, we have the following result:
\begin{Prop}\label{prop:ce}
Assume that each $(\bm x_n,y_n)$ is sampled from the joint distribution ${\cal P}_{{\cal X}{\cal Y}}$ uniformly at random. Let $\Delta^M$ denotes the probability simplex such that $\Delta^{M} = \{\bm u \in \mathbb{R}^M~| \bm u \ge \bm 0, \bm 1^{\top}\bm u=1\}$. Then, assuming $\bm A_m$'s are invertible, as $N$ grows to infinity, $\bm p_{\bm \theta}(\bm x_n) = \bm p(\bm x_n)$ and $\bm B_m = \bm A_m^{-1}, \forall m$, for any $\bm d \in \Delta^M$, the objective function in \eqref{eq:criterion_ce} attains the optimum value.
\end{Prop}
The proof of Proposition \ref{prop:ce} is given in supplementary material in Sec. \ref{app:prop1}.
Proposition \ref{prop:ce} suggests that the criterion in \eqref{eq:criterion_ce} attains the optimum value when the optimization variables $\bm B_m$'s are the inverses of the confusion matrices $\bm A_m$'s. Using the relation in \eqref{eq:pm}, these matrices can also be interpreted as the reverse confusion matrices with each entry indicating the conditional probability distribution ${\sf Pr}(y_n=k|\widehat{y}_n^{(m)}=k')$.

\noindent
{\bf Subspace distance-based regularization loss.} Here, we design a regularization loss to achieve the second objective $\textit{(ii)}$.
As we discussed, the model in \eqref{eq:subspace} implies that the pseudo-label vector $\bm p(\bm x_n)$ of the ID data lie in the subspace spanned by the column vectors of $\bm W$, i.e., ${\cal R}(\bm W)$. Note that the dimension of this low-dimensional subspace is $K$.  Consequently, there is an associated null space ${\cal N}(\bm W)$  that lies orthogonal to ${\cal R}(\bm W)$  and is defined as
${\cal N}(\bm W) = \{\bm p \in \mathbb{R}^{MK} : \bm W^{\top} \bm p =\bm 0\}$,
which is of dimension $(M-1)K$. 
Our idea is to ensure that the learned feature vector $\bm p_{\bm \theta}(\bm x)$ of any ID sample $\bm x$ lies in the $K$-dimensional subspace ${\cal R}(\bm W)$. In this way,  The vector $\bm p_{\bm \theta}(\bm x^*)$ for any OOD sample $\bm x^*$ will most likely lie orthogonally in the null space--see Fig. \ref{fig:ood:a}. This occurs because OOD samples do not belong to any ID classes and therefore fail to conform to the low-dimensional subspace structure described in \eqref{eq:pm}.   Note that having 
$M>1$ is important here; otherwise, when $M=1$, the null space will be empty (containing only zero vector), making it difficult to distinguish between ID and OOD samples.

Leveraging this idea, we proceed to maximize the projection of the vector $\bm p_{\bm \theta}(\bm x_n)$ to the range space ${\cal R}(\bm W)$ during training. Alternatively, this can be achieved by minimizing the projection of the vector $\bm p_{\bm \theta}(\bm x_n)$ to the null space ${\cal N}(\bm W)$.  Hence, we propose the following regularization loss:
\begin{align} \label{eq:reg}
    {\sf L}_{reg}(\{\bm x_n\}_{n=1}^N, \bm W) = \sum_{n=1}^N\frac{\|{\sf Proj}_{\bm W^{\perp}}(\bm p_{\bm \theta}(\bm x_n))\|_2}{\|\bm p_{\bm \theta}(\bm x_n)\|_2},
\end{align}
where ${\sf Proj}_{\bm W^{\perp}}$ denotes the projection to the null space ${\cal N}(\bm W)$ and is given by \cite{boyd2018introduction}:
\begin{align*}
    {\sf Proj}_{\bm W^{\perp}}(\bm p_{\bm \theta}(\bm x_n)) = (\bm I- \bm W \left( \bm W^{\top} \bm W \right)^{-1} \bm W^{\top}) \bm p_{\bm \theta}(\bm x_n).
\end{align*}
Thus, our final learning criterion is given by
\begin{align*}
    {\rm minimize}&~ {\sf L}_{CE} +\lambda {\sf L}_{reg}(\{\bm x_n\}_{n=1}^N, \bm W)\\
    {\rm subject~to}&~\bm B = [d_1\bm B_1, d_2 \bm B_2, \dots, d_M\bm B_M], \bm W = [\bm B^{-\top}_1,  \bm B^{-\top}_2, \dots, \bm B^{-\top}_M]^{\top}\\
    &~\bm d \ge \bm 0,~ \bm 1^{\top}\bm d=1, ~\bm 1^{\top} \bm B_m = \bm 1^{\top}, \bm B_m \ge \bm 0,
\end{align*}
where ${\sf L}_{CE}$ is given by cross-entropy loss in \eqref{eq:objective} and $\bm B^{-\top}_m$ denotes the transpose taken to the inverse of the matrix $\bm B_m$.
As we analyzed in Proposition \ref{prop:ce}, the matrices $\bm B_m^{-1}$'s are expected to learn the confusion matrices $\bm A_m$'s under optimal settings, thereby learning the subspace as represented by \eqref{eq:subspace}. The proposed learning framework is illustrated in Fig. \ref{fig:PRISM} and
we name the framework as PRISM, that stands for {\bf \underline{P}}seudo-label {\bf \underline{R}}epresentation {\bf \underline{I}}nduced {\bf \underline{S}}ubspace {\bf \underline{M}}odeling for OOD detection.

\section{Experiments}






 In this section, we showcase the effectiveness of our proposed OOD detection framework.
\paragraph{Datasets.}
During training, we use CIFAR-10 and CIFAR-100~\cite{Krizhevsky2009LearningML} as ID datasets. CIFAR-10 consists of 50,000 training images and 10,000 testing images, with 10 classes. CIFAR-100 contains the same number of training and testing images, but is divided into 100 fine-grained classes, making classification as well as OOD detection more challenging. During the test time, we evaluate our method with several OOD datasets such as SVHN \cite{netzer}, FashionMNIST \cite{XiaoFahionmnist}, LSUN \cite{yu2016lsun}, iSUN \cite{panisun}, Texture \cite{Cimpoi}, and Places365 \cite{zhou2016}.

\begin{table*}[t] 
\centering

\renewcommand{\arraystretch}{1.2} 
\LARGE 
\setlength{\tabcolsep}{4pt} 

\resizebox{\textwidth}{!}{ 
\begin{tabular}{l|cc|cc|cc|cc|cc|cc||cc}
\hline
\textbf{Method} & \multicolumn{2}{c|}{\textbf{SVHN}} & \multicolumn{2}{c|}{\textbf{FashionMNIST}} & \multicolumn{2}{c|}{\textbf{LSUN}} & \multicolumn{2}{c|}{\textbf{iSUN}} & \multicolumn{2}{c|}{\textbf{Texture}} & \multicolumn{2}{c||}{\textbf{Places365}} & \multicolumn{2}{c}{\textbf{Average}} \\ \hline
                & \textbf{FPR $\downarrow$} & \textbf{AUROC $\uparrow$} & \textbf{FPR $\downarrow$} & \textbf{AUROC $\uparrow$} & \textbf{FPR $\downarrow$} & \textbf{AUROC $\uparrow$} & \textbf{FPR $\downarrow$} & \textbf{AUROC $\uparrow$} & \textbf{FPR $\downarrow$} & \textbf{AUROC $\uparrow$} & \textbf{FPR $\downarrow$} & \textbf{AUROC $\uparrow$} & \textbf{FPR} $\downarrow$ & \textbf{AUROC}$\uparrow$ \\ \hline

MSP \cite{hendrycks2017baseline} & 68.52 & 88.38 & 43.08 & 94.00 & 25.28 & 96.70 & 43.55 & 94.29 & 66.03 & 87.26 & 63.60 & 88.23 & 51.68 & 91.81 \\  
ODIN \cite{liang2018enhancing} & 36.46 & 91.88 & 12.39 & 97.77 & \textbf{1.87} & \textbf{99.54} & \textbf{3.72} & \textbf{99.16} & 56.37 & 84.19 & \textbf{44.71} & {90.07} & 25.92 & 93.77 \\  
Energy Score~\cite{liu2020energy} & 53.72 & 90.65 & 13.02 & 97.64 & \textbf{2.12} & \textbf{99.44} & 9.34 & 98.15 & 61.01 & 85.93 & \textbf{43.58} & \textbf{90.73} & 30.46 & 93.76 \\  
ReAct \cite{sun2021react} & 93.30 & 72.29 & 28.45 & 95.50 & 13.12 & 97.54 & 43.12 & 92.59 & 87.57 & 67.16 & 57.24 & 86.79 & 53.80 & 85.31 \\  
Mahalanobis \cite{lee2018simple} & 14.70 & 95.98 & 69.80 & 79.71 & 50.18 & 88.50 & 29.79 & 90.91 & 28.44 & 89.40 & 90.40 & 52.94 & 47.22 & 82.91 \\  
KNN+~\cite{sun2022deepnn} & 6.43 & 98.86 & 13.72 & 97.59 & 6.97 & 98.67 & 10.60 & 98.13 & 21.54 & 95.99 & 53.70 & 87.73 & 18.83 & 96.16 \\  
CIDER~\cite{ming2023hypersphericalood}    & 43.71 & 92.77  & 51.12  & 89.93 & 84.39 &  80.08  & 98.81 & 60.09  & 85.94  & 70.64 &  88.22 & 75.01  & 75.37 &78.08 \\ 
SSD+~\cite{SehwagSSD}    & 60.96 & 86.73  & 91.74  & 78.94 & 40.27  &  89.85  & 91.50 & 71.33  & 57.85  & 77.94 & 94.76 & 49.81  & 72.85 &75.77 \\

SNN\cite{ghosal2023overcomeood} & 2.67 & 99.52 & \textbf{11.16} & \textbf{98.05} & 5.19 & 99.13 & \textbf{8.74} & \textbf{98.44} & 19.84 & \textbf{96.51} & 45.49 & \textbf{90.35} & 15.85 & 96.67 \\  
\hline
\textbf{PRISM} ($\lambda =0$)  & \textbf{2.02}  & \textbf{99.63} & 12.12 & 97.45 & 4.01 & 99.22 & 10.07  & 98.16 & \textbf{19.81} & 96.28  & 51.79 & 88.07  & 16.97 & 96.47 \\ 
\textbf{PRISM}  & \textbf{1.64} & \textbf{99.58} & \textbf{11.47} & \textbf{97.89} & 4.26 & 99.18 & 10.48 & 98.07 & \textbf{15.73} & \textbf{97.09} & 44.94 & 89.45  & \textbf{14.75} & \textbf{96.88} \\  
\hline
\end{tabular}
 }
\vspace{3pt}
\caption{OOD detection performance using different OOD datasets on CIFAR-10.} 
\label{tab:table-1}
\end{table*}

\paragraph{Baselines.}
We compare our method with several existing baselines. Specifically, we consider MSP \cite{hendrycks2017baseline}, ODIN \cite{liang2018enhancing}, Energy Score \cite{Liu2020EnergybasedOD}, ReAct \cite{sun2021react}, Mahalanobis \cite{lee2018simple}, KNN+ \cite{sun2022deepnn}, CIDER \cite{ming2023hypersphericalood}, SSD+ \cite{SehwagSSD}, and SNN \cite{ghosal2023overcomeood}. MSP, ODIN and Energy Score are softmax-based approaches.
 MSP relies only on softmax output of the model, while ODIN uses an additional temperature scaling hyperparameter. Energy Score computes an energy-based metric from the model outputs, identifying test samples with higher energy as OOD. ReAct is a logit-based approach. Mahalanobis, KNN+, CIDER, SSD+, and SNN are distance-based approaches. For MSP, ODIN, Energy Score, ReAct, and SNN, the deep neural network encoder is trained using the standard cross-entropy loss. For KNN+ and SSD+, supervised contrastive loss \cite{khosla2020supervised} is used. CIDER is trained using a maximum likelihood estimation-based loss together with dispersion regularization. 


\paragraph{OOD detection score.} \label{sec:score}
During testing, we employ a non-parametric $k$NN-based scoring on the penultimate feature space, as we have observed a strong ID-OOD separation using our approach---see Fig. \ref{fig:ood:b}, where the ID-OOD separability is enhanced quite significantly in the penultimate feature space. We follow a similar strategy used by the recent distance-based methods \cite{sun2022deepnn,ghosal2023overcomeood,ming2023hypersphericalood} to formulate the $k$NN-based score. Specifically, we estimate the Euclidean distance between the penultimate features of the training samples, \( \bm{h}(\bm{x}_n) \), and that of the test sample, i.e., \( \bm h(\bm{x}^*)\). 
We also perform $\ell_2$-normalization on the $\bm h$ vectors of both training and test data. For e.g., for a given test sample $\bm x^*$, we compute \( \bm u^* = \bm{h}(\bm{x}^*) / \|\bm{h}(\bm{x}^*)\|_2 \), ensuring scale invariance across samples. The OOD detection score is then formulated as $s(\bm x^*) = - \|\bm u^* - \bm u_k\|_2$ and the detection output $\bm g_{\tau}(\bm x^*)$ is as given in \eqref{eq:detect}. Here
\( \bm u_k \) represents the $k$th nearest neighbour embedding from the training data, and \( \tau \) is a threshold chosen to classify a high fraction (e.g., 95\%) of ID data correctly. Note that due to our proposed training strategy, these embeddings are learned, leveraging the pseudo-label subspace-based representation  so that ID-OOD separability is effectively ensured. 




\paragraph{Evaluation metrics.}
We evaluate our OOD detection performance using three widely used metrics. First, FPR@95 measures the proportion of OOD samples that are incorrectly classified as ID when the true positive rate is set to 95\%. Lower values of FPR@95 indicate better performance. Next, we consider AUROC, which represents the area under the receiver operating characteristic (ROC) curve. Finally, we report ID Accuracy (ID ACC), which measures the classification accuracy on ID data.

\begin{table*}[t]
\centering
\renewcommand{\arraystretch}{1.2} 
\LARGE
\setlength{\tabcolsep}{4pt} 

\resizebox{\textwidth}{!}{
\begin{tabular}{l|cc|cc|cc|cc|cc|cc||cc}
\hline
\textbf{Method} & \multicolumn{2}{c|}{\textbf{SVHN}} & \multicolumn{2}{c|}{\textbf{FashionMNIST}} & \multicolumn{2}{c|}{\textbf{LSUN}} & \multicolumn{2}{c|}{\textbf{iSUN}} & \multicolumn{2}{c|}{\textbf{Texture}} & \multicolumn{2}{c||}{\textbf{Places365}} & \multicolumn{2}{c}{\textbf{Average}} \\ \hline
                & \textbf{FPR $\downarrow$} & \textbf{AUROC $\uparrow$} & \textbf{FPR $\downarrow$} & \textbf{AUROC $\uparrow$} & \textbf{FPR $\downarrow$} & \textbf{AUROC $\uparrow$} & \textbf{FPR $\downarrow$} & \textbf{AUROC $\uparrow$} & \textbf{FPR $\downarrow$} & \textbf{AUROC $\uparrow$} & \textbf{FPR $\downarrow$} & \textbf{AUROC $\uparrow$} & \textbf{FPR} $\downarrow$ & \textbf{AUROC} $\uparrow$\\ \hline

MSP \cite{hendrycks2017baseline} & 96.22 & 50.50 & 68.24 & 84.82 & 66.86 & 80.44 & 89.44 & 72.09 & 98.07 & 45.14 & 92.38 & 63.57 & 85.87 & 66.43 \\  
ODIN \cite{liang2018enhancing} & 97.51 & 35.60 & 44.94 & 90.48 & 51.15 & 87.60 & 74.84 & 81.09 & 98.37 & 35.12 & 91.99 & 61.80 & 76.80 & 64.28 \\  
Energy Score \cite{liu2020energy} & 92.89 & 74.39 & 18.07 & 96.53 & \textbf{11.77} & \textbf{97.49} & 71.12 & 82.77 & 93.10 & 62.12 & 82.09 & 74.65 & 61.84 & 81.99 \\  
ReAct \cite{sun2021react} & 99.28 & 27.50 & 72.98 & 83.60 & 62.30 & 83.04 & 93.39 & 69.31 & 98.60 & 29.85 & 92.34 & 60.38 & 86.15 & 58.94 \\  
Mahalanobis \cite{lee2018simple} & 43.87 & 89.90 & 99.17 & 43.37 & 96.32 & 46.96 & 37.01 & 92.13 & 34.52 & 89.22 & 96.00 & 52.50 & 67.81 & 68.68 \\  
KNN+ \cite{sun2022deepnn} & 20.71 & 96.46 & \textbf{11.33} & \textbf{97.64} & \textbf{21.87} & \textbf{94.37} & 48.42 & 87.11 & 27.34 & 93.32 & 90.53 & 63.63 & 36.70 & 88.76 \\ 
CIDER~\cite{ming2023hypersphericalood}   &  99.66 & 51.06 &99.90 &  27.38 & 99.31 & 9.42  &  99.84 & 27.25 & 93.72 & 39.31  & 99.81 & 24.37 & 98.71 &29.80  \\ 
SSD+~\cite{SehwagSSD}    &  99.41 & 47.89 &97.76 &  52.44 & 93.88 & 63.20  &  98.14 & 49.47 & 77.20 & 60.70  & 98.63 & 40.81  & 94.17 &52.41  \\ 
SNN \cite{ghosal2023overcomeood} & \textbf{14.95} & \textbf{97.16} & 24.68 & 95.53 & 26.00 & 93.85 & 45.37 & 87.43 & \textbf{24.04} & \textbf{94.63} & \textbf{85.24} & \textbf{68.01} & 36.05 & 89.77 \\  
\hline
\textbf{PRISM} ($\lambda =0$)  & \textbf{10.18}  & \textbf{97.93} & 18.98 & 96.78 & 39.43 & 90.43 & \textbf{42.41}  & \textbf{90.60} & \textbf{25.85}  & \textbf{94.37}  & \textbf{85.22} & \textbf{69.64}  & 37.68 & 89.96 \\ 
\textbf{PRISM}  & 15.96 & 96.91 & \textbf{14.45} & \textbf{97.15} & 30.98 & 92.49 & \textbf{26.77} &  \textbf{94.27} & 26.86 & 94.09 & 86.60 & 67.15  & \textbf{33.60} & \textbf{90.34} \\ 
\hline
\end{tabular}
}
\vspace{3pt}
\caption{OOD detection performance using different OOD datasets on CIFAR-100.} 
\label{tab:table-2}
\end{table*}

\begin{figure*}[t]
    \centering
    \begin{subfigure}[t]{0.48\linewidth}
        \centering
        \includegraphics[width=\linewidth]{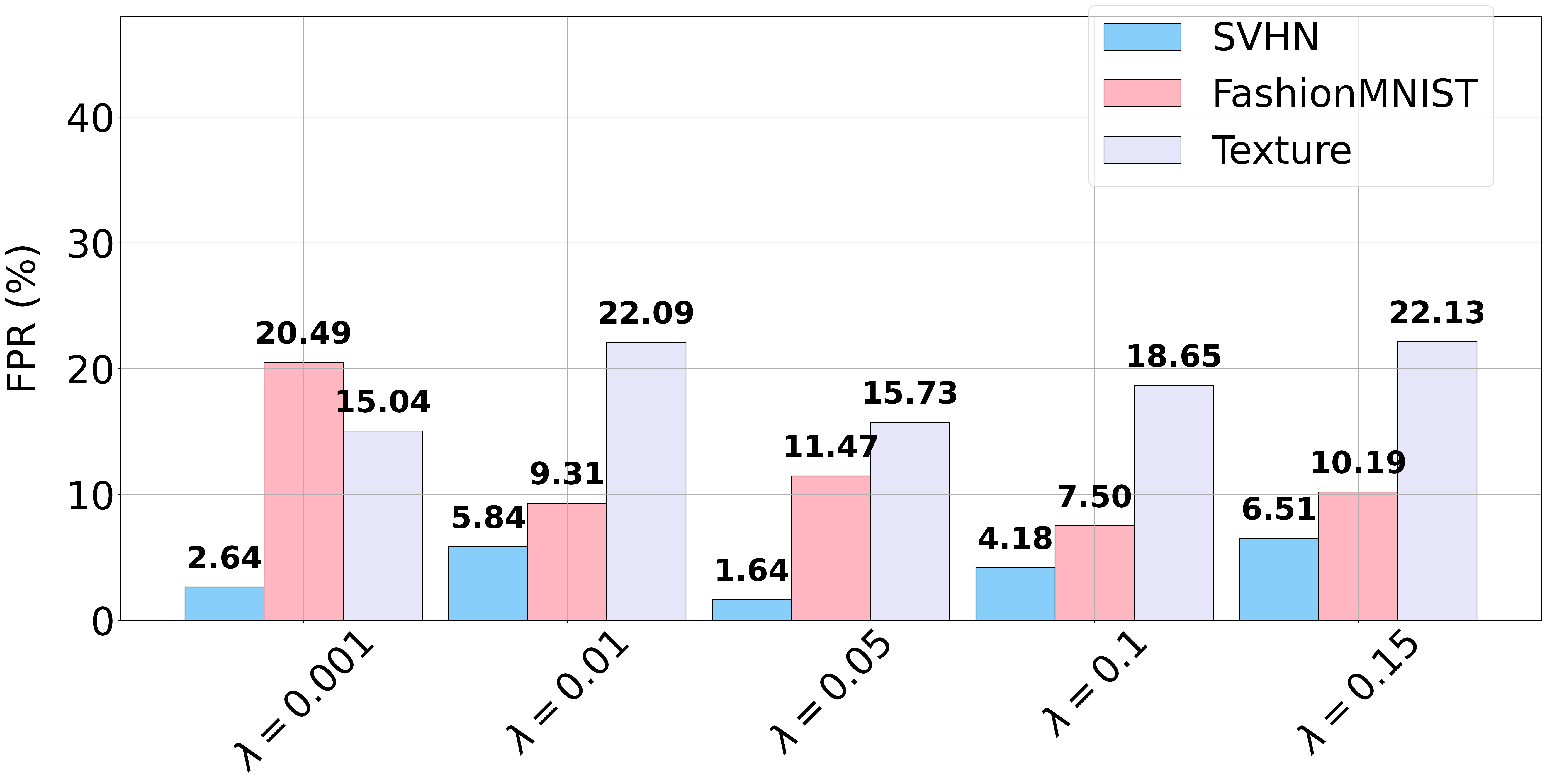} 
        \caption{}
        \label{fig:lamda_1}
    \end{subfigure}
    \hfill
        \begin{subfigure}[t]{0.48\linewidth}
        \centering
        \includegraphics[width=0.8\linewidth]{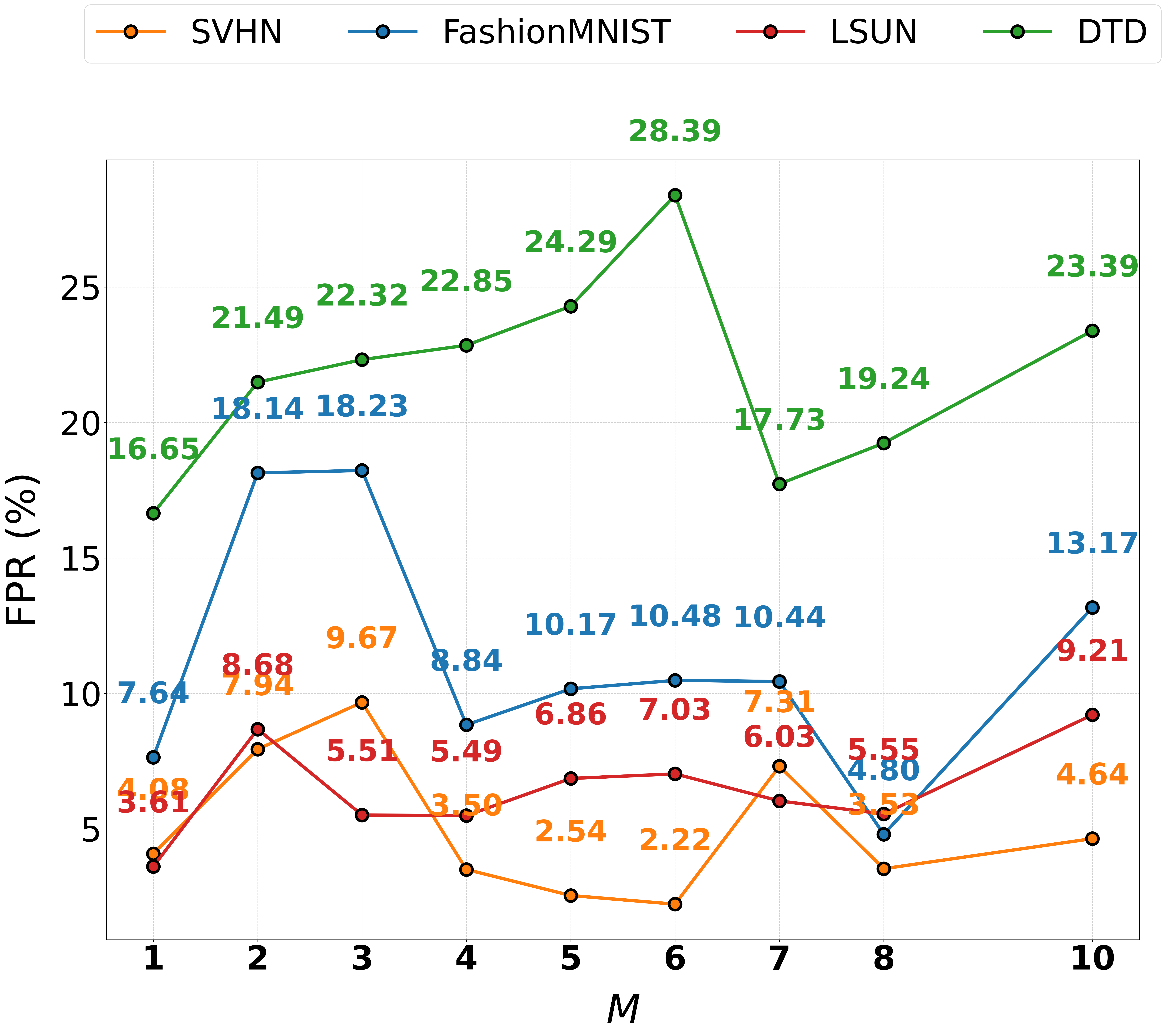}
        \caption{}
        \label{fig:ood_m_values}
    \end{subfigure}
    \caption{ (a) FPR performance of PRISM for different $\lambda$ values and (b) OOD detection performance of PRISM for different $M$. CIFAR-10 is the ID dataset. The encoder model is DenseNet-101.
    }
    \label{fig:lamda_study}
\end{figure*}
\begin{figure*}[t]
    \centering
    \begin{subfigure}[t]{0.39\linewidth}
        \centering
    \includegraphics[width=\linewidth]{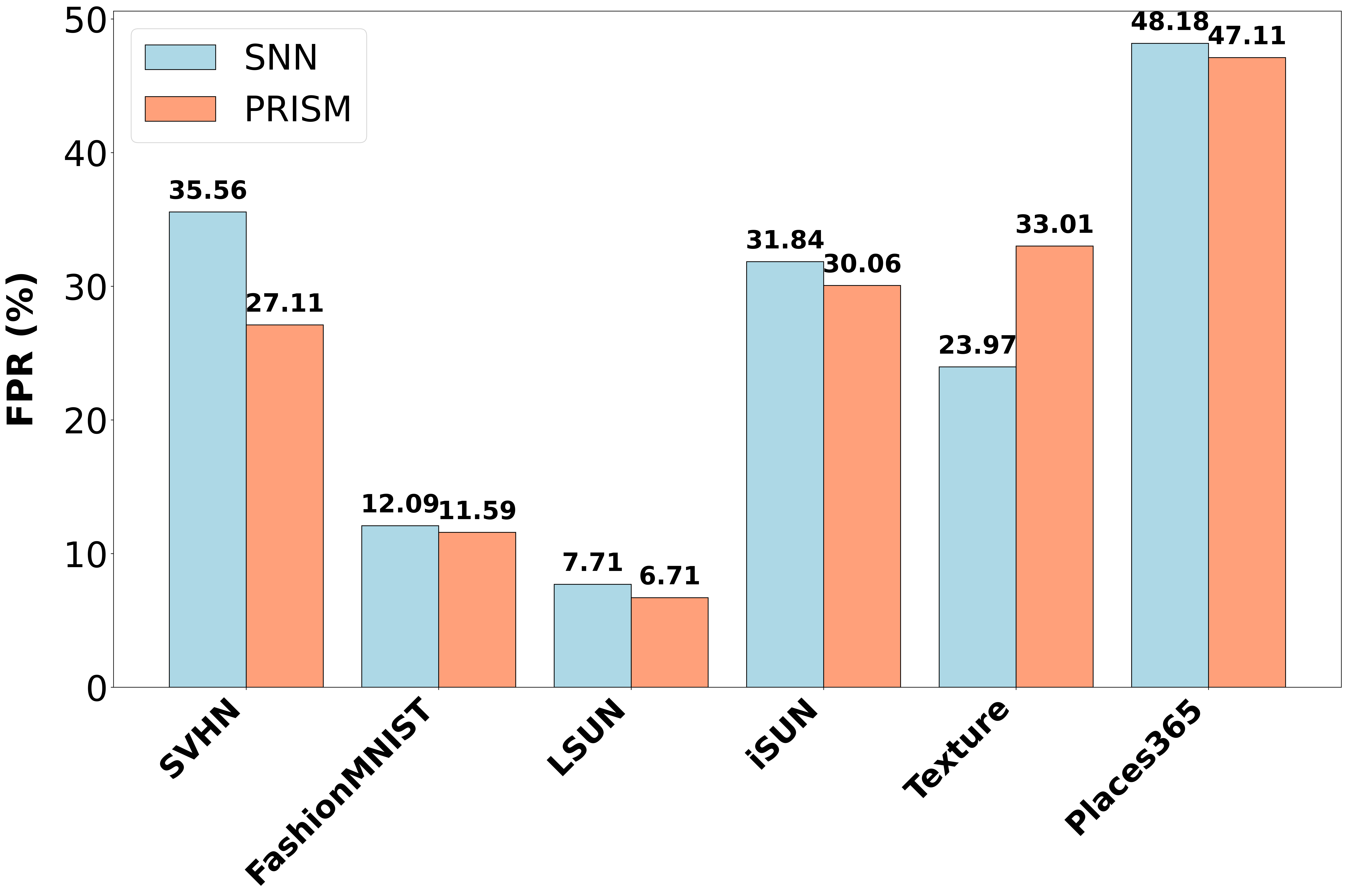}
      \caption{}
       \label{fig:resnet_plot}
    \end{subfigure}
    \hfill
    \begin{subfigure}[t]{0.59\linewidth}
        \centering
    \centering
    \includegraphics[width=1\linewidth]{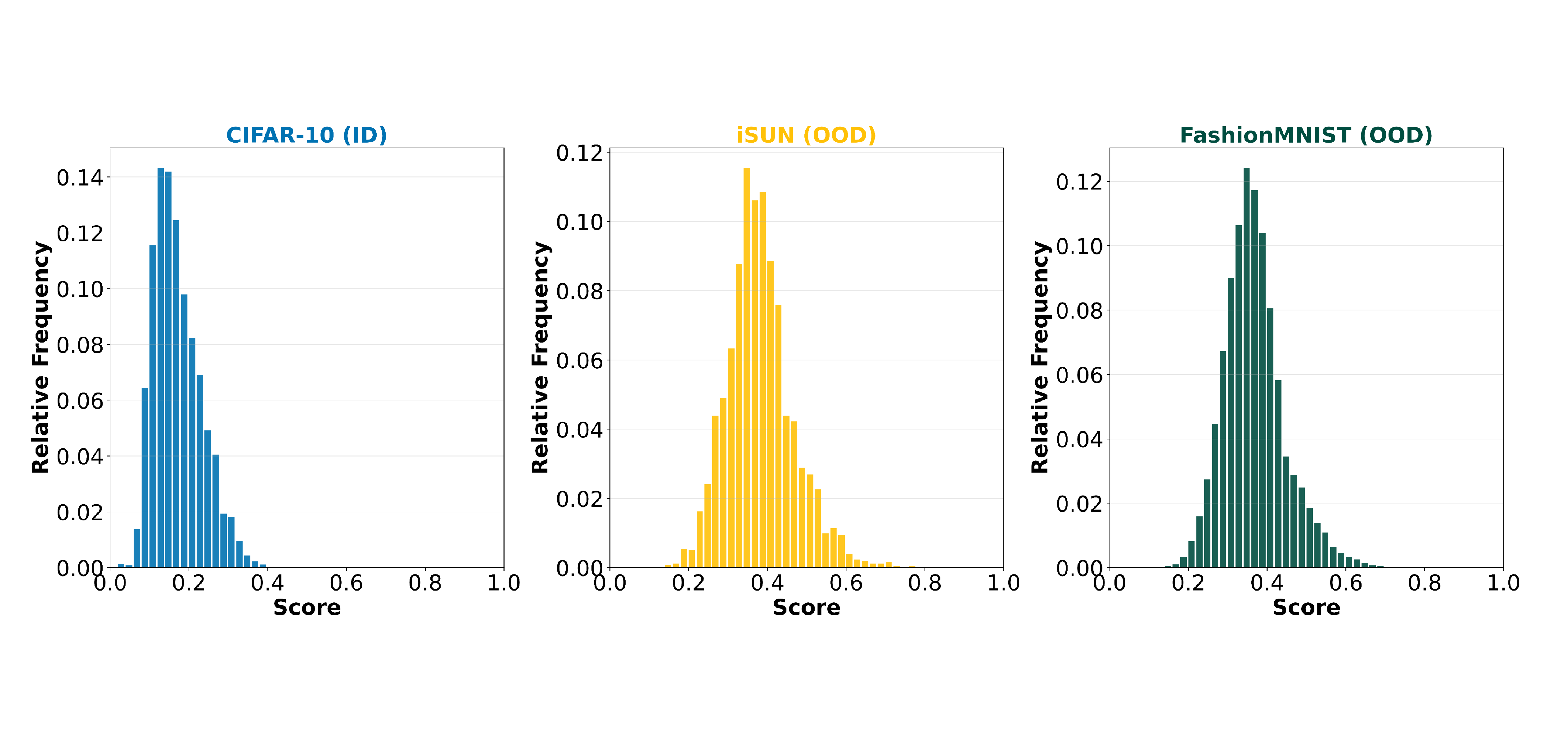}

      \caption{}
      \label{fig:knn_histogram}
    \end{subfigure}
    \caption{(a) OOD detection performance of PRISM using ResNet-50 encoder model across different OOD datasets. (b) OOD detection score distribution for CIFAR-10, iSUN, and FashionMNIST datasets. CIFAR-10 is the ID dataset, while others are OOD datasets. }
    \label{fig:bar_plot_resnet-M}
\end{figure*}
    

    

\paragraph{Experiment settings.}
We use a CNN-based architecture, DenseNet-101\cite{densenet}, as the backbone model for training on CIFAR-10 and CIFAR-100 datasets. We train the model from scratch using our ID data. During training for CIFAR-10, we set the number of epochs to 100 and use a batch size of 64. First, we transform the penultimate layer’s features, which is of dimension $L=342$, into $MK$ dimensional representation, and then we apply a softmax to each $K$-sized block of the transformed features---see Fig. \ref{fig:PRISM}. For CIFAR-10, we choose stochastic gradient descent (SGD) as the optimizer with a momentum of 0.9 and a weight decay of $1\times10^{-4}$.  The weight matrices $\bm B_m$'s initialized as identity matrices of size $K \times K$ and $\bm d$ vectors are initialized to uniform values, i.e., each $d_m= 1/M$. 
We tune the hyperparameters $\lambda$ from the set of values $\{0.001, 0.005, 0.01, 0.05, 0.1, 0.15\}$ and the number of pseudo labels $M$ from 1 to 10.
For CIFAR-100, the SGD optimizer is utilized to train the neural network parameters $\bm \theta$ with a learning rate of 0.2, momentum of 0.9, and a weight decay of $1\times 10^{-4}$. To train the weight matrices $\bm B_m$'s and $\bm d$, we use the Adam optimizer with a learning rate of 0.2, and a weight decay of $1\times 10^{-6}$. 
For the regularization loss, we use the Neumann series approximation \cite{NeumannNewton1877} to efficiently perform the inverse of the $\bm B_m$ matrices. Although the $\bm B_m$ matrices introduce additional computational overhead during training, it remains minimal as long as $M$ is kept small (as we will see, smaller $M$ is sufficient across several scenarios).  
We follow the same strategy for initialization and for tuning the hyperparameters $\lambda$  and $M$, as in the case of CIFAR-10.

 \paragraph{Results.}
Table \ref{tab:table-1} presents the OOD detection performance of various baselines and our method PRISM on the CIFAR-10 dataset. All methods are trained using the DenseNet-101 architecture on the ID dataset, i.e., CIFAR-10, without access to any OOD datasets. The results indicate that PRISM consistently achieves strong performance across multiple OOD datasets.
For the three challenging OOD datasets, SVHN, FashionMNIST, and Texture (where OOD samples appear semantically similar to ID samples, also referred to as near-OOD setting), the best baseline SNN attains an average FPR of approximately 11.22\% and an average AUROC of 98.03\%, whereas PRISM improves these metrics to 9.61\% and 98.19\%, respectively. One can also note that the proposed subspace distance-based regularization brings substantial improvement in OOD detection, especially in challenging datasets like SVHN, FashionMNIST, and Texture. While ODIN and Energy Score perform well on LSUN and iSUN, they struggle with more challenging near-OOD cases like SVHN and Texture, exhibiting significantly worse performance compared to PRISM.

Table \ref{tab:table-2} compares OOD detection performance of the baselines and our method using CIFAR-100 as the ID dataset with the DenseNet-101 model. One can note that PRISM maintains robust performance across different OOD datasets. 
Compared to the competing baseline SNN, PRISM achieves a lower average FPR (33.60\% vs. 36.05\%) and an improved average AUROC (90.34\% vs. 89.77\%), with particularly strong detection performance on the FashionMNIST, iSUN, and LSUN datasets.
Other baselines like ODIN and Energy Score struggle significantly in CIFAR-100 scenario, which is indeed more challenging than CIFAR-10 setting. Overall, PRISM outperforms all baselines on average for both CIFAR-10 and CIFAR-100 (see the last column of Table \ref{tab:table-1} and \ref{tab:table-2}), highlighting the robustness of our framework for OOD detection. We also analyze the ID classification accuracy of our method and observes that PRISM performs comparably to existing baselines in both CIFAR-10 and CIFAR-100—--see the results in Table~\ref{tab:table-sup-5} of the supplementary material. 






\paragraph{Ablation study.}

\noindent
\textit{ (i) Different $\lambda$ values.} We analyze the effect of the regularization hyperparameter $\lambda$ on the OOD performance of our method. Fig. \ref{fig:lamda_1} presents the FPR metric on the SVHN, FashionMNIST, and Texture datasets for various $\lambda$ values. The results indicate that $\lambda = 0.05$ yields the best FPR performance across all three datasets. It can be also noted that very small as well as relatively large $\lambda$ values degrade FPR performance. 
More detailed results, including additional datasets and ID accuracies, are provided in the supplementary material in Table \ref{tab:table-sup-2}.

\noindent
\textit{(ii) Different $M$ values.} Here, we study the effect of varying $M$ on the OOD performance. 
We use the CIFAR-10 as ID dataset and use the DenseNet-101 encoder architecture. We keep all other settings same as before, except $M$. Figure \ref{fig:ood_m_values} shows the FPR values on the SVHN, FashionMNIST, LSUN, and Texture datasets for $M$ varying from 1 to 10. We observe that increasing $M$ from 2 to 4 lowers the FPR values. With $M=5$ and $M=6$, the FPR continues to decrease, but increasing $M$ to 7 or higher begins to worsen performance. Larger settings like $M=10$ further degrade the performance, which is expected as it involves learning more number of parameters for the confusion matrices. 
More detailed analysis related to varying $M$ can be found in the supplementary material in Table \ref{tab:table-sup-3}.

\noindent
\textit{(iii) Different encoder architectures.} We also evaluate PRISM with an alternative encoder architecture, namely ResNet-50 \cite{he2016deep}. Figure \ref{fig:resnet_plot} compares the FPR performance of PRISM against the best performing baseline method SNN on the ResNet-50 architecture. One can note that our method shows consistently better OOD detection performance in this setting as well. 

\noindent
{\it (iv) Histogram of detection scores.}
 In Fig. \ref{fig:knn_histogram}, we present the histogram of the $k$NN-based detection scores, as explained in Sec. \ref{sec:score}, for CIFAR-10, iSUN, and FashionMNIST datasets. The plot demonstrates a clear distinction between ID and OOD scores, with smaller scores for CIFAR-10 dataset and larger scores for the OOD datasets iSUN and FashionMNIST. 
 More histogram results are presented in the supplementary section, where we also analyze the histogram of the regularization scores (i.e., \eqref{eq:reg}) for both ID and OOD datasets.

 
 %



\section{Related Work}

Our work advances the state-of-the-art in distance-based OOD detection. We devote this section to provide a detailed overview of the distance-based approaches that inspired developing our idea.
This line of methods rely on the intuition that ID samples cluster more tightly in feature space, whereas OOD samples lie farther from the feature space. 
For example, the method in \cite{mahalanobis2018generalized} utilizes the Mahalanobis distance score, that models ID data features distributed as Gaussian distribution and measures the Mahalanobis distance between a test sample and the closest class-conditional Gaussian distribution to detect OOD samples. Variants of  Mahalanobis distance-based methods  are also proposed, where \cite{lee2018simple} incorporates features from multiple early layers to improve robustness, and \cite{ren2021mahalanobis} introduces class-specific Mahalanobis distance scores for better detection. \cite{hendrycks2022icml} proposed a KL divergence-based score by computing the smallest KL divergence between a test sample’s softmax distribution and the mean softmax distribution of each ID class.
 The $k$-NN-based approach in \cite{sun2022deepnn} evaluates OOD scores by computing the Euclidean distance between a test sample’s feature representation and its nearest ID neighbors. Similarly, the work in \cite{ghosal2023overcomeood} extends the $k$NN-based approach by projecting features into a lower-dimensional subspace while preserving the most informative parts. The work in \cite{SehwagSSD} used supervised contrastive loss \cite{khosla2020supervised} to learn feature representations, which are then used with Mahalanobis distance or kNN distance for OOD detection. A recent approach \cite{ming2023hypersphericalood} assumes that the feature embeddings are hyperspherical in nature, following a von Mises-Fisher distribution \cite{mardia2000directional} and learn them by jointly optimizing an inter-class dispersion loss and intra-class compactness loss. 
More discussion about related works, e.g., softmax-based and logit-based methods, are presented in the supplementary section in Sec. \ref{app:related}.
\section{Conclusion}
In this work, we introduced a novel framework for OOD detection that overcomes the limitations of the existing feature-based methods relying on strong distributional assumptions and only high dimensional feature representations. By leveraging pseudo-label-induced subspace representations and a carefully designed learning criterion, we provide a more flexible and effective approach to improving ID-OOD separability. 
Our extensive experimental results demonstrate the superiority of our method across multiple benchmarks and challenging OOD scenarios.

{
    \bibliographystyle{unsrt}
    \bibliography{main}

\begin{thebibliography}{10}

\bibitem{goodfellow2014explaining}
Ian~J Goodfellow, Jonathon Shlens, and Christian Szegedy.
\newblock Explaining and harnessing adversarial examples.
\newblock {\em arXiv preprint arXiv:1412.6572}, 2014.

\bibitem{geiger2012we}
Andreas Geiger, Philip Lenz, and Raquel Urtasun.
\newblock Are we ready for autonomous driving? the kitti vision benchmark suite.
\newblock In {\em 2012 IEEE conference on computer vision and pattern recognition}, pages 3354--3361. IEEE, 2012.

\bibitem{thomas2017unsupervised}
Thomas Schlegl, Philipp Seeb{\"o}ck, Sebastian~M. Waldstein, Ursula Schmidt-Erfurth, and Georg Langs.
\newblock Unsupervised anomaly detection with generative adversarial networks to guide marker discovery.
\newblock In {\em Information Processing in Medical Imaging}, pages 146--157. Springer International Publishing, 2017.

\bibitem{hendrycks2016baseline}
Dan Hendrycks and Kevin Gimpel.
\newblock A baseline for detecting misclassified and out-of-distribution examples in neural networks.
\newblock {\em arXiv preprint arXiv:1610.02136}, 2016.

\bibitem{yang2024generalized}
Jingkang Yang, Kaiyang Zhou, Yixuan Li, and Ziwei Liu.
\newblock Generalized out-of-distribution detection: A survey.
\newblock {\em International Journal of Computer Vision}, 132(12):5635--5662, 2024.

\bibitem{liang2018enhancing}
Shiyu Liang, Yixuan Li, and R.~Srikant.
\newblock Enhancing the reliability of out-of-distribution image detection in neural networks.
\newblock In {\em International Conference on Learning Representations (ICLR)}, 2018.

\bibitem{hendrycks2022scaling}
Dan Hendrycks, Steven Basart, Mantas Mazeika, Andy Zou, Joseph Kwon, Mohammadreza Mostajabi, Jacob Steinhardt, and Dawn Song.
\newblock Scaling out-of-distribution detection for real-world settings.
\newblock In {\em Proceedings of the 39th International Conference on Machine Learning (ICML)}, pages 8759--8773. PMLR, 2022.

\bibitem{sun2022dice}
Yiyou Sun and Yixuan Li.
\newblock Dice: Leveraging sparsification for out-of-distribution detection.
\newblock In {\em Computer Vision – ECCV 2022}, pages 691--708. Springer Nature Switzerland, 2022.

\bibitem{sun2021react}
Yiyou Sun, Chuan Guo, and Yixuan Li.
\newblock React: Out-of-distribution detection with rectified activations.
\newblock In {\em Advances in Neural Information Processing Systems (NeurIPS)}, pages 144--157. Curran Associates, Inc., 2021.

\bibitem{dong2022neural}
Xin Dong, Junfeng Guo, Ang Li, Wei-Te Ting, Cong Liu, and HT~Kung.
\newblock Neural mean discrepancy for efficient out-of-distribution detection.
\newblock In {\em Proceedings of the IEEE/CVF Conference on Computer Vision and Pattern Recognition}, pages 19217--19227, 2022.

\bibitem{lee2018simple}
Kimin Lee, Kibok Lee, Honglak Lee, and Jinwoo Shin.
\newblock A simple unified framework for detecting out-of-distribution samples and adversarial attacks.
\newblock In {\em Advances in Neural Information Processing Systems (NeurIPS)}. Curran Associates, Inc., 2018.

\bibitem{sun2022deepnn}
Yiyou Sun, Yifei Ming, Xiaojin Zhu, and Yixuan Li.
\newblock Out-of-distribution detection with deep nearest neighbors.
\newblock In {\em Proceedings of the 39th International Conference on Machine Learning (ICML)}, pages 20827--20840. PMLR, 2022.

\bibitem{ming2023hypersphericalood}
Yifei Ming, Yiyou Sun, Ousmane Dia, and Yixuan Li.
\newblock How to exploit hyperspherical embeddings for out-of-distribution detection?, 2023.

\bibitem{SehwagSSD}
Vikash Sehwag, Mung Chiang, and Prateek Mittal.
\newblock {SSD:} {A} unified framework for self-supervised outlier detection.
\newblock {\em CoRR}, abs/2103.12051, 2021.

\bibitem{ghosal2023overcomeood}
Soumya~Suvra Ghosal, Yiyou Sun, and Yixuan Li.
\newblock How to overcome curse-of-dimensionality for out-of-distribution detection?
\newblock {\em Proceedings of the AAAI Conference on Artificial Intelligence}, 38(18):19849--19857, Mar. 2024.

\bibitem{mahalanobis2018generalized}
Prasanta~Chandra Mahalanobis.
\newblock On the generalized distance in statistics.
\newblock {\em Sankhy{\=a}: The Indian Journal of Statistics, Series A (2008-)}, 80:S1--S7, 2018.

\bibitem{khosla2020supervised}
Prannay Khosla, Piotr Teterwak, Chen Wang, Aaron Sarna, Yonglong Tian, Phillip Isola, Aaron Maschinot, Ce~Liu, and Dilip Krishnan.
\newblock Supervised contrastive learning.
\newblock {\em Advances in neural information processing systems}, 33:18661--18673, 2020.

\bibitem{mcinnes2018umap}
Leland McInnes, John Healy, and James Melville.
\newblock Umap: Uniform manifold approximation and projection for dimension reduction.
\newblock {\em arXiv preprint arXiv:1802.03426}, 2018.

\bibitem{li2021provably}
Xuefeng Li, Tongliang Liu, Bo~Han, Gang Niu, and Masashi Sugiyama.
\newblock Provably end-to-end label-noise learning without anchor points.
\newblock In {\em Proceedings of International Conference on Machine Learning}, pages 6403--6413, 2021.

\bibitem{tanno2019learning}
Ryutaro Tanno, Ardavan Saeedi, Swami Sankaranarayanan, Daniel~C Alexander, and Nathan Silberman.
\newblock Learning from noisy labels by regularized estimation of annotator confusion.
\newblock In {\em Proceedings of the IEEE/CVF conference on computer vision and pattern recognition}, pages 11244--11253, 2019.

\bibitem{ibrahim2023deep}
Shahana Ibrahim, Tri Nguyen, and Xiao Fu.
\newblock Deep learning from crowdsourced labels: Coupled cross-entropy minimization, identifiability, and regularization.
\newblock In {\em Proceedings of International Conference on Learning Representations}, 2023.

\bibitem{dawid1979maximum}
Alexander~Philip Dawid and Allan~M Skene.
\newblock Maximum likelihood estimation of observer error-rates using the {EM} algorithm.
\newblock {\em Applied statistics}, pages 20--28, 1979.

\bibitem{boyd2018introduction}
Stephen Boyd and Lieven Vandenberghe.
\newblock {\em Introduction to Applied Linear Algebra: Vectors, Matrices, and Least Squares}.
\newblock Cambridge University Press, 2018.

\bibitem{Krizhevsky2009LearningML}
Alex Krizhevsky.
\newblock Learning multiple layers of features from tiny images.
\newblock 2009.

\bibitem{netzer}
Yuval Netzer, Tao Wang, Adam Coates, Alessandro Bissacco, Bo~Wu, and Andrew~Y. Ng.
\newblock Reading digits in natural images with unsupervised feature learning.
\newblock In {\em NIPS Workshop on Deep Learning and Unsupervised Feature Learning 2011}, 2011.

\bibitem{XiaoFahionmnist}
Han Xiao, Kashif Rasul, and Roland Vollgraf.
\newblock Fashion-mnist: a novel image dataset for benchmarking machine learning algorithms.
\newblock {\em CoRR}, abs/1708.07747, 2017.

\bibitem{yu2016lsun}
Fisher Yu, Ari Seff, Yinda Zhang, Shuran Song, Thomas Funkhouser, and Jianxiong Xiao.
\newblock Lsun: Construction of a large-scale image dataset using deep learning with humans in the loop, 2016.

\bibitem{panisun}
Junting Pan and Xavier Gir{\'{o}}{-}i{-}Nieto.
\newblock End-to-end convolutional network for saliency prediction.
\newblock {\em CoRR}, abs/1507.01422, 2015.

\bibitem{Cimpoi}
Mircea Cimpoi, Subhransu Maji, Iasonas Kokkinos, Sammy Mohamed, and Andrea Vedaldi.
\newblock Describing textures in the wild.
\newblock {\em CoRR}, abs/1311.3618, 2013.

\bibitem{zhou2016}
Bolei Zhou, Aditya Khosla, {\`{A}}gata Lapedriza, Antonio Torralba, and Aude Oliva.
\newblock Places: An image database for deep scene understanding.
\newblock {\em CoRR}, abs/1610.02055, 2016.

\bibitem{hendrycks2017baseline}
Dan Hendrycks and Kevin Gimpel.
\newblock A baseline for detecting misclassified and out-of-distribution examples in neural networks.
\newblock In {\em 5th International Conference on Learning Representations, {ICLR} 2017, Toulon, France, April 24-26, 2017, Conference Track Proceedings}. OpenReview.net, 2017.

\bibitem{liu2020energy}
Weitang Liu, Xiaoyun Wang, John Owens, and Yixuan Li.
\newblock Energy-based out-of-distribution detection.
\newblock In {\em Advances in Neural Information Processing Systems (NeurIPS)}, pages 21464--21475. Curran Associates, Inc., 2020.

\bibitem{Liu2020EnergybasedOD}
Weitang Liu, Xiaoyun Wang, John~Douglas Owens, and Yixuan Li.
\newblock Energy-based out-of-distribution detection.
\newblock {\em ArXiv}, abs/2010.03759, 2020.

\bibitem{densenet}
Gao Huang, Zhuang Liu, and Kilian~Q. Weinberger.
\newblock Densely connected convolutional networks.
\newblock {\em CoRR}, abs/1608.06993, 2016.

\bibitem{NeumannNewton1877}
C.~Neumann and I.~Newton.
\newblock {\em Untersuchungen über das logarithmische und Newton’sche Potential}.
\newblock B. G. Teubner, 1877.

\bibitem{he2016deep}
Kaiming He, Xiangyu Zhang, Shaoqing Ren, and Jian Sun.
\newblock Deep residual learning for image recognition.
\newblock In {\em Proceedings of the IEEE Conference on Computer Vision and Pattern Recognition (CVPR)}, pages 770--778, 2016.

\bibitem{ren2021mahalanobis}
Jie Ren, Stanislav Fort, Jeremiah Liu, Abhijit~Guha Roy, Shreyas Padhy, and Balaji Lakshminarayanan.
\newblock A simple fix to mahalanobis distance for improving near-ood detection.
\newblock 2021.

\bibitem{hendrycks2022icml}
Dan Hendrycks, Steven Basart, Mantas Mazeika, Andy Zou, Joseph Kwon, Mohammadreza Mostajabi, Jacob Steinhardt, and Dawn Song.
\newblock Scaling out-of-distribution detection for real-world settings.
\newblock In {\em Proceedings of the 39th International Conference on Machine Learning (ICML)}, pages 8759--8773. PMLR, 2022.

\bibitem{mardia2000directional}
Kanti~V. Mardia, Peter~E. Jupp, and K.V. Mardia.
\newblock {\em Directional Statistics, Volume 2}.
\newblock Wiley Online Library, 2000.

\bibitem{guo2017calibration}
Chuan Guo, Geoff Pleiss, Yu~Sun, and Kilian~Q. Weinberger.
\newblock On calibration of modern neural networks.
\newblock In {\em Proceedings of the 34th International Conference on Machine Learning (ICML)}, pages 1321--1330. PMLR, 2017.

\bibitem{LiuGenEntropy}
Xixi Liu, Yaroslava Lochman, and Christopher Zach.
\newblock Gen: Pushing the limits of softmax-based out-of-distribution detection.
\newblock In {\em 2023 IEEE/CVF Conference on Computer Vision and Pattern Recognition (CVPR)}, pages 23946--23955, 2023.

\bibitem{song2022rankfeat}
Yue Song, Nicu Sebe, and Wei Wang.
\newblock Rankfeat: Rank-1 feature removal for out-of-distribution detection.
\newblock In {\em Advances in Neural Information Processing Systems (NeurIPS)}, 2022.

\bibitem{Ammar2023NECONC}
Mouin~Ben Ammar, Nacim Belkhir, Sebastian Popescu, Antoine Manzanera, and Gianni Franchi.
\newblock Neco: Neural collapse based out-of-distribution detection.
\newblock {\em ArXiv}, abs/2310.06823, 2023.

\bibitem{Cook2020OutlierDT}
Matthew Cook, Alina Zare, and Paul~D. Gader.
\newblock Outlier detection through null space analysis of neural networks.
\newblock {\em ArXiv}, abs/2007.01263, 2020.

\bibitem{Wang2022ViMOW}
Haoqi Wang, Zhizhong Li, Litong Feng, and Wayne Zhang.
\newblock Vim: Out-of-distribution with virtual-logit matching.
\newblock {\em 2022 IEEE/CVF Conference on Computer Vision and Pattern Recognition (CVPR)}, pages 4911--4920, 2022.

\end{thebibliography}
}
  \onecolumn
\newpage
\appendix
\begin{center}
    \normalsize
    {\bf Supplementary Material of ``Pseudo-label Induced Subspace Representation Learning for Robust Out-of-Distribution Detection''}
\end{center}

\section{Notation} \label{app:notation}
 We use the following notation throughout the paper: $x$, $\x$, $\X$, and $\tX$ represent a scalar, a vector, a matrix, and a tensor, respectively. Both $x_i$ and $[\bm x]_i$ denote the $i$th entry of the vector $\x$.  $[\bm X]_{i,j}$ denote the $(i,j)$th entry of the matrix $\bm X$. $\bm x_i$ denotes the $i$th row of the matrix $\bm X$;
$[I]$ means an integer set $\{1,2,\ldots,I\}$. 
  $^\T$  denote transpose.
  $\bm X \ge \bm 0$ implies that all the entries of the matrix $\bm X$ are non-negative. $\mathbb{I}[A]$ denotes an indicator function for the event $A$ such that $\mathbb{I}[A]=1$ if the event $A$ happens, otherwise $\mathbb{I}[A]=0$. ${\sf CE}(\bm x,y) = -\sum_{k=1}^K \mathbb{I}[{y}=k]\log(\bm x(k))$ denotes the cross entropy function. 
  $\bm I$ denotes an identity matrix of appropriate size. $\bm 1_K$ denotes an all-one vector of size $K$. $|{\cal C}|$ denotes the cardinality of the set ${\cal C}.$ $\Delta^{K}$ denotes a $(K-1)$-dimensional probability simplex such that
     $\Delta^{K} = \{\bm u \in \mathbb{R}^K~| \bm u \ge \bm 0, \bm 1^{\top}\bm u=1\}$.

\section{Proof of Proposition \ref{prop:ce}} \label{app:prop1}

Given the conditions in Proposition \ref{prop:ce}, i.e., each $(\bm x_n,y_n)$ is sampled from the joint distribution ${\cal P}_{{\cal X}{\cal Y}}$ uniformly at random. and  $N$ grows to infinity, minimizing the cross-entropy term in \eqref{eq:criterion_ce} is equivalent to minimizing the Kullback-Leibler (KL) divergence between $\bm f(\bm x_n)$ and $\bm B \bm p_{\theta}(\bm x_n)$ for each $n$. Then, the objective function attains the optimal value, when
\begin{align}\label{eq:Bp}
     \bm B \bm p_{\theta}(\bm x_n) = \bm f(\bm x_n),
\end{align}
where


\begin{align*}
    \bm B &= [d_1\bm B_1, \dots, d_M\bm B_M]\\
    \bm p_{\bm \theta}(\bm x_n)  &= \begin{bmatrix}\bm p_{\bm \theta}^{(1)}(\bm x_n) \\ \vdots \\ \bm p_{\bm \theta}^{(M)}(\bm x_n)\end{bmatrix}.
\end{align*}

Now, let us consider the term on the left hand side of \eqref{eq:Bp}, i.e.,
\begin{align*}
    \bm B\bm p_{\bm \theta}(\bm x_n) &= d_1\bm B_1 \bm p_{\bm \theta}^{(1)}(\bm x_n)+ d_2\bm B_2 \bm p_{\bm \theta}^{(2)}(\bm x_n) + \dots+ d_M\bm B_M \bm p_{\bm \theta}^{(M)}(\bm x_n).
\end{align*}

By substituting $\bm B_m= \bm A_m^{-1}$ and $\bm p_{\bm \theta}^{(m)}(\bm x_n) = \bm p_m(\bm x_n)= \bm A_m \bm f(\bm x_n)$, for any $\bm d \in \Delta^M$, we have
\begin{align*}
    \bm B\bm p_{\bm \theta}(\bm x_n) &= d_1\bm B_1 \bm p_{\bm \theta}^{(1)}(\bm x_n)+ d_2\bm B_2 \bm p_{\bm \theta}^{(2)}(\bm x_n) + \dots+ d_M\bm B_M \bm p_{\bm \theta}^{(M)}(\bm x_n)\\
    &= d_1\bm A_1^{-1} (\bm A_1 \bm f(\bm x_n))+ d_2\bm A_2^{-1} (\bm A_2 \bm f(\bm x_n)) + \dots+ d_M\bm A_M^{-1} (\bm A_M \bm f(\bm x_n))\\
    &= d_1 \bm f(\bm x_n) + d_2 \bm f(\bm x_n)+\dots+d_M \bm f(\bm x_n)\\
    &=\bm f(\bm x_n),
\end{align*}
where the last equality used the constraint that $\bm d \in \Delta^M$ and hence $\sum_{m=1}^M d_m=1$. This implies that the objective function reaches the optimal values using the given values of $\bm B_m$ and $\bm p_{\bm \theta}(\bm x_n)$.

\section{More Discussion on Related Works} \label{app:related}

\noindent
{\bf Softmax-based approaches.}
Softmax-based OOD detection methods leverage the intuition that ID samples exhibit higher confidence scores and lower entropy compared to OOD samples. The maximum softmax probability method in~\cite{hendrycks2017baseline} determines the OOD score by taking the highest softmax probability among the predicted class probabilities. The work in \cite{guo2017calibration} introduces a temperature scaling parameter to better enhance ID-OOD differentiation. ODIN \cite{liang2018enhancing} further enhances this approach by applying small input perturbations and combining them with temperature scaling, leading to improved OOD detection performance. Another approach, named generalized entropy \cite{LiuGenEntropy}, considers the entire predictive distribution rather than just the maximum softmax probability. It quantifies how much the predicted distribution deviates from a one-hot distribution and uses this to measure the OODness of the samples.


\noindent
{\bf Logit-based scoring.}
Instead of computing softmax probabilities, logit-based methods directly analyze the raw logits (pre-softmax outputs) to derive OOD scores. The maximum logit score method~\cite{hendrycks2022scaling} selects the highest logit value as the OOD score.
Several approaches refine this idea by computing energy-based \cite{liu2020energy} scores from the logit output. For example, ReAct \cite{sun2021react} proposes clipping activation values to mitigate the influence of high-magnitude outliers, while RankFeat \cite{song2022rankfeat} refines activations by subtracting the rank of the feature matrix to provide improved OOD discrimination.
Another method, DICE~\cite{sun2022dice}, introduces weight sparsification to identify the most influential features for OOD detection. 

\par
Other approaches in OOD detection include leveraging the complex properties of deep neural network representations. For example, \cite{Ammar2023NECONC} propose a post-hoc method that exploits neural collapse phenomena to distinguish OOD samples by projecting in-distribution features onto a structured principal space. In a related approach, \cite{Cook2020OutlierDT} integrate outlier detection directly into the classification pipeline through a null space analysis, allowing the network to simultaneously maintain high classification performance and identify anomalies. Complementing these methods, \cite{Wang2022ViMOW} introduce an approach, that fuses class-agnostic residual information with traditional logits via virtual-logit matching to enhance the accuracy of OOD detection.


\section{Additional Experiment Results}
\textbf{Different Initialization Study}. Here, we analyze the effect of different initialization for the optimization variables, specifically for $\bm B$ and $\bm d$. We perform two different initializations on the confusion matrix $\bm{B}$: random initialization and identity matrix initialization. For both versions of $\bm{B}$, we vary the $\bm{d}$ values in three different ways. First, we set $\bm{d}$ with uniform values, i.e., $d_m = 1/M, \forall m$ and keep without updating during optimization. Next, we set $\bm{d}$ with uniform values, but allow the optimizer to update them. Finally, we add a linear layer to transform the penultimate layer feature values to match the dimension of $\bm{d}$, and assign those values in $\bm{d}$ to induce some instance-dependence. The results corresponding to these settings are provided in Table \ref{tab:table-sup-1}.

\begin{table*}[t]
\centering
\renewcommand{\arraystretch}{1.2} 
\setlength{\tabcolsep}{5pt} 
\small 

\resizebox{\textwidth}{!}{
\begin{tabular}{l|cc|cc|cc|cc|cc|cc|c}
\hline
\textbf{Initialization Method} & \multicolumn{2}{c|}{\textbf{SVHN}} & \multicolumn{2}{c|}{\textbf{FashionMNIST}} & \multicolumn{2}{c|}{\textbf{LSUN}} & \multicolumn{2}{c|}{\textbf{iSUN}} & \multicolumn{2}{c|}{\textbf{DTD}} & \multicolumn{2}{c|}{\textbf{Places365}} & \\ \hline
 & \textbf{FPR $\downarrow$} & \textbf{AUROC $\uparrow$} & \textbf{FPR $\downarrow$} & \textbf{AUROC $\uparrow$} & \textbf{FPR $\downarrow$} & \textbf{AUROC $\uparrow$} & \textbf{FPR $\downarrow$} & \textbf{AUROC $\uparrow$} & \textbf{FPR $\downarrow$} & \textbf{AUROC $\uparrow$} & \textbf{FPR $\downarrow$} & \textbf{AUROC $\uparrow$} &  \textbf{ID ACC} $\uparrow$ \\ \hline

\multicolumn{14}{l}{\textbf{Without Regularization ($\lambda = 0$)}} \\ \hline
Random $\bm B_m$, random $\bm d$ & 2.02 & 99.63 & 12.12 & 97.45 & 4.01 & 99.22 & 10.07 & 98.16 & 19.81 & 96.28 & 51.79 & 88.07 & 90.03 \\
Identity $\bm B_m$, random $\bm d$  & 3.55 & 99.34 & 11.94 & 97.77 & 4.17 & 99.20 & 11.99 & 97.76 & 19.40 & 96.53 & 50.75 & 88.41 & 89.71 \\
Identity $\bm B_m$, uniform fixed $\bm d$ & 26.92 & 96.02 & 13.88 & 97.32 & 10.25 & 98.08 & 15.70 & 96.98 & 27.41 & 94.79 & 49.09 & 87.97 & 90.75 \\
Identity $\bm B_m$, uniform but learnable $\bm d$  & 2.57 & 99.48 & 8.89 & 98.28 & 3.42 & 99.34 & 5.50 & 98.97 & 21.77 & 95.71 & 43.80 & 90.20 & 76.83 \\
Identity $\bm B_m$, linearly reduced $\bm d$  & 3.41 & 99.32 & 16.99 & 97.04 & 5.96 & 98.34 & 9.65 & 98.19 & 14.66 & 97.43 & 51.95 & 86.86 & 89.42 \\ \hline

\multicolumn{14}{l}{\textbf{With Regularization ($\lambda \neq 0$)}} \\ \hline
Identity $\bm B_m$, uniform fixed  $\bm d$ & 6.19 & 98.89 & 6.79 & 98.74 & 4.64 & 99.16 & 7.67 & 98.51 & 21.88 & 96.31 & 38.55 & 91.89 & 94.10 \\
Identity $\bm B_m$, uniform but learnable $\bm d$ & 2.54 & 99.51 & 10.17 & 98.17 & 6.86 & 98.73 & 12.03 & 97.82 & 24.29 & 95.18 & 43.48 & 90.63 & 94.53 \\
Identity $\bm B_m$, linearly reduced $\bm d$ & 2.80 & 99.34 & 5.53 & 98.91 & 3.16 & 99.36 & 15.43 & 97.29 & 16.81 & 96.96 & 42.34 & 90.19 & 94.27 \\ \hline
\end{tabular}
}
\vspace{3pt}
\caption{Comparison of different initialization methods for OOD detection performance using CIFAR-10 datasets. 
}
\label{tab:table-sup-1}
\end{table*}

\noindent
\textbf{Different \( \lambda \) values.}  
Table \ref{tab:table-sup-2} presents a detailed analysis of the ablation study on \( \lambda \). Here, we use identity initialization for $\bm B_m$ matrices and uniform initialization for $\bm d$ vector. We observe that lower values of \( \lambda \) negatively impact ID accuracy, while also leading to suboptimal FPR and AUROC across all datasets. As \( \lambda \) increases, performance improves, and we find that \( \lambda = 0.05 \) achieves the best balance in terms of FPR and AUROC. However, further increasing \( \lambda \) beyond this point tends to degrade overall performance. In Table \ref{tab:table-sup100-2}, we present similar analysis on CIFAR-100 dataset by varying $\lambda$.

\begin{table*}[!htbp]
    \centering
    
    \Large
    \setlength{\tabcolsep}{5pt} 
    \renewcommand{\arraystretch}{1.2} 
    \resizebox{\textwidth}{!}{%
    \begin{tabular}{|l|cc|cc|cc|cc|cc|cc|c|}
        \hline
        \textbf{$\lambda$} & \multicolumn{2}{c|}{\textbf{SVHN}} & \multicolumn{2}{c|}{\textbf{FashionMNIST}} & \multicolumn{2}{c|}{\textbf{LSUN}} & \multicolumn{2}{c|}{\textbf{iSUN}} & \multicolumn{2}{c|}{\textbf{Texture}} & \multicolumn{2}{c|}{\textbf{Places365}} &  \\ \hline
                & \textbf{FPR $\downarrow$} & \textbf{AUROC $\uparrow$} & \textbf{FPR $\downarrow$} & \textbf{AUROC $\uparrow$} & \textbf{FPR $\downarrow$} & \textbf{AUROC $\uparrow$} & \textbf{FPR $\downarrow$} & \textbf{AUROC $\uparrow$} & \textbf{FPR $\downarrow$} & \textbf{AUROC $\uparrow$} & \textbf{FPR $\downarrow$} & \textbf{AUROC $\uparrow$} & \textbf{ID ACC} $\uparrow$ \\ \hline
        0.001 & 2.64 & 99.48  & 20.49 & 96.01  & 6.94 & 98.77 &  14.60 & 97.42 & \textbf{15.04} & \textbf{97.37} &  55.58 & 85.96 &  87.69 \\
        0.01 & 5.84 & 98.82 & 9.31 & 98.15  & 5.42 & 99.04  & 13.77 & 97.65 & 22.09 & 96.24  & 44.71 & \textbf{90.84} & 94.28 \\
        0.05 & \textbf{1.64} & \textbf{99.58}  & 11.47 & 97.89 & \textbf{4.26} & \textbf{99.18}  & \textbf{10.48} & 98.07 & 15.73 & 97.09  & 44.94 & 89.45  & \textbf{94.55} \\
        0.1 & 4.18 & 99.25  & \textbf{7.50} & \textbf{98.48} &  6.32 & 98.70 & 13.41 & 97.60  & 18.65 & 96.55  & 46.04 & 89.77  & 94.31 \\
        0.15 & 6.51  & 97.24 & 10.19 & 98.32 & 5.76 & 98.90&\textbf{7.59} & \textbf{98.66}  & 22.13 & 95.39  & \textbf{40.79} & 91.21  & 94.01 \\
        \hline
    \end{tabular}%
    }
    \caption{OOD detection performance for different $\lambda$ values. ID dataset is CIFAR-10 and the encoder architecture is DenseNet-101. }
    \label{tab:table-sup-2}
\end{table*}


\begin{table*}[!htbp]
    \centering
    
    \Large
    \setlength{\tabcolsep}{5pt} 
    \renewcommand{\arraystretch}{1.2} 
    \resizebox{\textwidth}{!}{%
    \begin{tabular}{|l|cc|cc|cc|cc|cc|cc|c|}
        \hline
        \textbf{$\lambda$} & \multicolumn{2}{c|}{\textbf{SVHN}} & \multicolumn{2}{c|}{\textbf{FashionMNIST}} & \multicolumn{2}{c|}{\textbf{LSUN}} & \multicolumn{2}{c|}{\textbf{iSUN}} & \multicolumn{2}{c|}{\textbf{Texture}} & \multicolumn{2}{c|}{\textbf{Places365}} &  \\ \hline
                & \textbf{FPR $\downarrow$} & \textbf{AUROC $\uparrow$} & \textbf{FPR $\downarrow$} & \textbf{AUROC $\uparrow$} & \textbf{FPR $\downarrow$} & \textbf{AUROC $\uparrow$} & \textbf{FPR $\downarrow$} & \textbf{AUROC $\uparrow$} & \textbf{FPR $\downarrow$} & \textbf{AUROC $\uparrow$} & \textbf{FPR $\downarrow$} & \textbf{AUROC $\uparrow$} & \textbf{ID ACC} $\uparrow$ \\ \hline
        0.001 & 16.07 & \textbf{97.06}  & 18.30 & 96.63  & 40.90 & 89.94 &  34.10 & 92.12 & 33.44 & 92.63 &  89.33 & 66.91 &  \textbf{74.93} \\
         0.01& \textbf{15.96} & 96.91 & 14.45 & 97.15 & \textbf{30.98} & \textbf{92.49} & \textbf{26.77} &  \textbf{94.27} & \textbf{26.86} & \textbf{94.09} & \textbf{86.60} & \textbf{67.15}  & 74.83 \\ 
        0.05 & 18.12 & 96.18 & \textbf{10.90} & \textbf{97.77} & 39.73 & 88.44 & 27.69 &  93.68 & 29.29 & 93.51 & 88.61 & 66.81  & 74.63 \\  
        0.1 & 25.67 & 95.32  & 20.68 & 97.27 &  37.86 & 88.91 & 39.38 & 90.32  & 28.76 &  93.36  & 89.58 & 65.70  & 73.90 \\
        0.15 & 19.31  & 96.12 & 24.42 & 96.32 & 42.45 & 88.56 &34.78 & 93.67  & 31.84 & 92.89  & 88.76 & 65.94  & 74.20 \\
        \hline
    \end{tabular}%
    }
   \caption{OOD detection performance for different $\lambda$ values. ID dataset is CIFAR-100 and the encoder architecture is DenseNet-101. }
    \label{tab:table-sup100-2}
\end{table*}

\noindent
\textbf{Different \( M \) values.}  
Table \ref{tab:table-sup-3} presents a detailed analysis of the ablation study on \( M \). We observe similar trends in AUROC and FPR across different datasets. The results indicate that the optimal AUROC values are generally achieved when \( M = 5 \) or \( M = 6 \). Similarly to FPR, choosing a very low \( M \) does not yield favorable AUROC results for most datasets. Conversely, excessively high \( M \) values degrade performance in both FPR and AUROC. While some datasets achieve good FPR and AUROC for \( M = 1 \), lower \( M \) values reduce ID accuracy. Thus, selecting a moderate \( M \) provides an optimal balance across all datasets. Similar observations are noted in Table \ref{tab:table-sup-M100} for the CIFAR-100 dataset.


\begin{table*}[!htbp]
    \centering
    
    \Large
    \setlength{\tabcolsep}{5pt} 
    \renewcommand{\arraystretch}{1.2} 
    \resizebox{\textwidth}{!}{%
    \begin{tabular}{|l|cc|cc|cc|cc|cc|cc|c|}
        \hline
         & \multicolumn{2}{c|}{\textbf{SVHN}} & \multicolumn{2}{c|}{\textbf{FashionMNIST}} & \multicolumn{2}{c|}{\textbf{LSUN}} & \multicolumn{2}{c|}{\textbf{iSUN}} & \multicolumn{2}{c|}{\textbf{Texture}} & \multicolumn{2}{c|}{\textbf{Places365}} &  \\ \hline
                & \textbf{FPR $\downarrow$} & \textbf{AUROC $\uparrow$} & \textbf{FPR $\downarrow$} & \textbf{AUROC $\uparrow$} & \textbf{FPR $\downarrow$} & \textbf{AUROC $\uparrow$} & \textbf{FPR $\downarrow$} & \textbf{AUROC $\uparrow$} & \textbf{FPR $\downarrow$} & \textbf{AUROC $\uparrow$} & \textbf{FPR $\downarrow$} & \textbf{AUROC $\uparrow$} & \textbf{ID ACC} $\uparrow$ \\ \hline
        $M=1$  & 5.10 & 99.01 & 8.86 & 98.37 & \textbf{4.75} & \textbf{99.09} & \textbf{7.79} & \textbf{98.56} & 17.27 & 96.86 & 46.95 & 89.69 & 89.46 \\
        $M=2$  & 7.94 & 98.58 & 18.14 & 96.49 & 8.68 & 98.49 & 23.42 & 96.20 & 21.49 & 96.18 & 54.12 & 88.06 & 94.55 \\
        $M=3$  & 9.67 & 98.41 & 18.23 & 95.99 & 5.51 & 98.99 & 10.09 & 98.19 & 22.32 & 95.75 & \textbf{40.23} & \textbf{90.90} & 86.12 \\
        $M=4$  & 3.50 & 99.23 & 8.84 & 98.27 & 5.49 & 98.99 & 11.31 & 97.98 & 22.85 & 95.86 & 44.16 & 90.67 & 94.45 \\
        $M=5$   & 2.54 &99.51 & 10.17 & 98.17 & 6.86 & 98.73 & 12.03 & 97.82 & 24.29 & 95.18 & 43.48 & 90.63 & \textbf{94.53} \\
        $M=6$  & \textbf{2.22} & \textbf{99.54} & 10.48 & 98.06 & 7.03 & 98.68 & 23.69 & 96.02 & 28.39 & 94.60 & 46.63 & 89.67 & 94.45 \\
        $M=7$  & 7.31 & 98.68 & 10.44 & 98.02 & 6.03 & 98.85 & 10.09 & 98.08 & \textbf{17.73} & \textbf{96.93} & 47.47 & 88.03 & 94.13 \\
        $M=8$  & 3.53 & 99.32 & \textbf{4.80} & \textbf{99.06} & 5.55 & 99.03& 10.28 & 98.16 & 19.24 & 96.48 & 49.35 & 88.77 & 94.10 \\
        $M=9$  & 3.17 & 99.42 & 8.48 & 98.34 & 6.77 & 98.75& 11.51& 97.81 & 21.05 & 96.10 & 41.37 & 90.71 & 93.86 \\
       $M=10$  & 4.64 & 99.11 & 13.17 & 97.10 & 9.21 & 98.24 & 15.98 & 97.05 & 23.39 & 95.72 & 47.40 & 88.57 & 92.96 \\
        \hline
    \end{tabular}%
    }
    \caption{OOD detection performance for different $M$ values. ID dataset is CIFAR-10 and the encoder architecture is DenseNet-101.}
    \label{tab:table-sup-3}
\end{table*}

\begin{table*}[!htbp]
    \centering
    
    \Large
    \setlength{\tabcolsep}{5pt} 
    \renewcommand{\arraystretch}{1.2} 
    \resizebox{\textwidth}{!}{%
    \begin{tabular}{|l|cc|cc|cc|cc|cc|cc|c|}
        \hline
         & \multicolumn{2}{c|}{\textbf{SVHN}} & \multicolumn{2}{c|}{\textbf{FashionMNIST}} & \multicolumn{2}{c|}{\textbf{LSUN}} & \multicolumn{2}{c|}{\textbf{iSUN}} & \multicolumn{2}{c|}{\textbf{Texture}} & \multicolumn{2}{c|}{\textbf{Places365}} &  \\ \hline
                & \textbf{FPR $\downarrow$} & \textbf{AUROC $\uparrow$} & \textbf{FPR $\downarrow$} & \textbf{AUROC $\uparrow$} & \textbf{FPR $\downarrow$} & \textbf{AUROC $\uparrow$} & \textbf{FPR $\downarrow$} & \textbf{AUROC $\uparrow$} & \textbf{FPR $\downarrow$} & \textbf{AUROC $\uparrow$} & \textbf{FPR $\downarrow$} & \textbf{AUROC $\uparrow$} & \textbf{ID ACC} $\uparrow$ \\ \hline
        $M=1$  & 19.43 & 96.42 & 14.63 & \textbf{97.39} & \textbf{24.71} & \textbf{94.64}  & 39.08 &91.89& 30.37 & 93.05 & 90.49 & 66.21 & 74.35  \\
        $M=2$  & \textbf{15.96} & \textbf{96.91} & \textbf{14.45} & 97.15 & 30.98 & 92.49 & \textbf{26.77} &  \textbf{94.27} & \textbf{26.86} & \textbf{94.09} & \textbf{86.60} & \textbf{67.15} & \textbf{74.83}\\
        $M=3$  & 18.33 & 96.38 &15.94 & 96.97 & 39.23 & 89.92  & 65.17  & 78.67 & 32.20 & 92.35 & 89.91 & 64.93 & 73.56 \\
        \hline
    \end{tabular}%
    }
    \caption{OOD detection performance for different $M$ values. ID dataset is CIFAR-100 and the encoder architecture is DenseNet-101.}
    \label{tab:table-sup-M100}
\end{table*}
\begin{table*}[!htbp] 
    \centering
    \Large
    \setlength{\tabcolsep}{5pt} 
    \renewcommand{\arraystretch}{1.2} 
    \resizebox{\textwidth}{!}{
    \begin{tabular}{l|cc|cc|cc|cc|cc|cc}
        \hline
        \textbf{Method} & \multicolumn{2}{c|}{\textbf{SVHN}} & \multicolumn{2}{c|}{\textbf{FashionMNIST}} & \multicolumn{2}{c|}{\textbf{LSUN}} & \multicolumn{2}{c|}{\textbf{iSUN}} & \multicolumn{2}{c|}{\textbf{DTD}} & \multicolumn{2}{c}{\textbf{Places365}} \\ \hline
        & \textbf{FPR $\downarrow$} & \textbf{AUROC $\uparrow$} & \textbf{FPR $\downarrow$} & \textbf{AUROC $\uparrow$} & \textbf{FPR $\downarrow$} & \textbf{AUROC $\uparrow$} & \textbf{FPR $\downarrow$} & \textbf{AUROC $\uparrow$} & \textbf{FPR $\downarrow$} & \textbf{AUROC $\uparrow$} & \textbf{FPR $\downarrow$} & \textbf{AUROC $\uparrow$} \\ \hline
        \textbf{SNN} & 35.56 & 93.26 & 12.09 & \textbf{98.07} & 7.71 & 98.77 & 31.84 & 94.61 & \textbf{23.97} & \textbf{95.64} & 48.18 & 90.50  \\ 
        \textbf{PRISM} & \textbf{27.11} & \textbf{95.78} & \textbf{11.59} & 97.85 & \textbf{6.71} & \textbf{98.58} & \textbf{30.06} & \textbf{94.68} & 33.01 & 93.43 & \textbf{47.11} & \textbf{89.22}  \\ \hline
    \end{tabular}
    }
    \caption{OOD detection performance comparison for different architecture. The ID dataset is CIFAR-10, and the encoder architecture is ResNet-50.}
    \label{tab:table-sup-4}
\end{table*}

\begin{table*}[!htbp] 
    \centering
    \Large
    \setlength{\tabcolsep}{5pt} 
    \renewcommand{\arraystretch}{1.2} 
    \resizebox{\textwidth}{!}{
    \begin{tabular}{l|cc|cc|cc|cc|cc|cc}
        \hline
        \textbf{Method} & \multicolumn{2}{c|}{\textbf{SVHN}} & \multicolumn{2}{c|}{\textbf{FashionMNIST}} & \multicolumn{2}{c|}{\textbf{LSUN}} & \multicolumn{2}{c|}{\textbf{iSUN}} & \multicolumn{2}{c|}{\textbf{DTD}} & \multicolumn{2}{c}{\textbf{Places365}} \\ \hline
        & \textbf{FPR $\downarrow$} & \textbf{AUROC $\uparrow$} & \textbf{FPR $\downarrow$} & \textbf{AUROC $\uparrow$} & \textbf{FPR $\downarrow$} & \textbf{AUROC $\uparrow$} & \textbf{FPR $\downarrow$} & \textbf{AUROC $\uparrow$} & \textbf{FPR $\downarrow$} & \textbf{AUROC $\uparrow$} & \textbf{FPR $\downarrow$} & \textbf{AUROC $\uparrow$} \\ \hline
        \textbf{SNN} & \textbf{17.24} & \textbf{96.47} & 19.21 & 96.01 & 54.34 & 83.29 & \textbf{59.85} & 81.94 & \textbf{31.24} & \textbf{93.39} & 85.67 & \textbf{68.19} \\
        \textbf{PRISM} & 22.11 & 96.18 & \textbf{17.59} & \textbf{97.15} & \textbf{46.71} & \textbf{89.58} & 62.06 & \textbf{82.68} & 33.01 & 92.93 & \textbf{85.11} & 66.22  \\ \hline
    \end{tabular}
    }
    \caption{OOD detection performance comparison for different architecture. The ID dataset is CIFAR-100, and the encoder architecture is ResNet-50.}
    \label{tab:table-cifar100}
\end{table*}

\begin{table}[t]
\centering
\renewcommand{\arraystretch}{1.1}
\small 
\setlength{\tabcolsep}{5pt} 
\begin{tabular}{l|cc}
\hline
\textbf{Method} & \textbf{CIFAR-10} & \textbf{CIFAR-100}  \\ \hline
MSP\cite{hendrycks2017baseline} & 94.03 & 75.21 \\  
ODIN\cite{liang2018enhancing} & 93.65 & 75.09 \\  
Energy Score\cite{liu2020energy} & 94.03 & 75.15 \\  
ReAct\cite{sun2021react} & 93.27 & 74.89 \\  
Mahalanobis\cite{lee2018simple} & 93.65 & 75.18 \\  
KNN+\cite{sun2022deepnn}  & 94.03 & 74.98 \\  
CIDER\cite{ming2023hypersphericalood} & 94.03 & 75.21 \\ 
SSD+\cite{SehwagSSD} & 94.03 & 75.19 \\ 
SNN\cite{ghosal2023overcomeood} & \textbf{94.15} & \textbf{75.21} \\  
PRISM ($\lambda =0$) & 90.03 & \textbf{75.33} \\  
PRISM & \textbf{94.55} & 74.83 \\  
\hline
\end{tabular}
\vspace{3pt}
\caption{ID classification accuracy (ID ACC $\uparrow$) for different OOD detection methods on CIFAR-10 and CIFAR-100.} 
\label{tab:table-sup-5}
\end{table}

\noindent
\textbf{Different Architecture} Table \ref{tab:table-sup-4} presents  detailed performance results for a different architecture, specifically ResNet-50, on CIFAR-10. We observe that PRISM outperforms the baseline SNN on most datasets. PRISM achieves a lower average FPR (25.93\%) compared to SNN (26.56\%), indicating improved OOD detection. Additionally, PRISM attains a competitive AUROC (94.92\%) close to SNN (95.14\%), further demonstrating its effectiveness as a robust OOD detection model. Table \ref{tab:table-cifar100} presents the OOD detection performance with CIFAR-100 using ResNet-50 model. The results also indicate that PRISM achieves high AUROC and low FPR in multiple OOD scenarios.


\noindent
\textbf{ID Classification Accuracy} We present the ID classification accuracy for both CIFAR-10 and CIFAR-100 in Table \ref{tab:table-sup-5}. For both datasets, our method PRISM performs better or is comparable to other baselines.


\noindent

    

\begin{figure*}[htbp]
    \centering
    \begin{subfigure}[t]{0.48\linewidth}
        \centering
        \includegraphics[width=\linewidth]{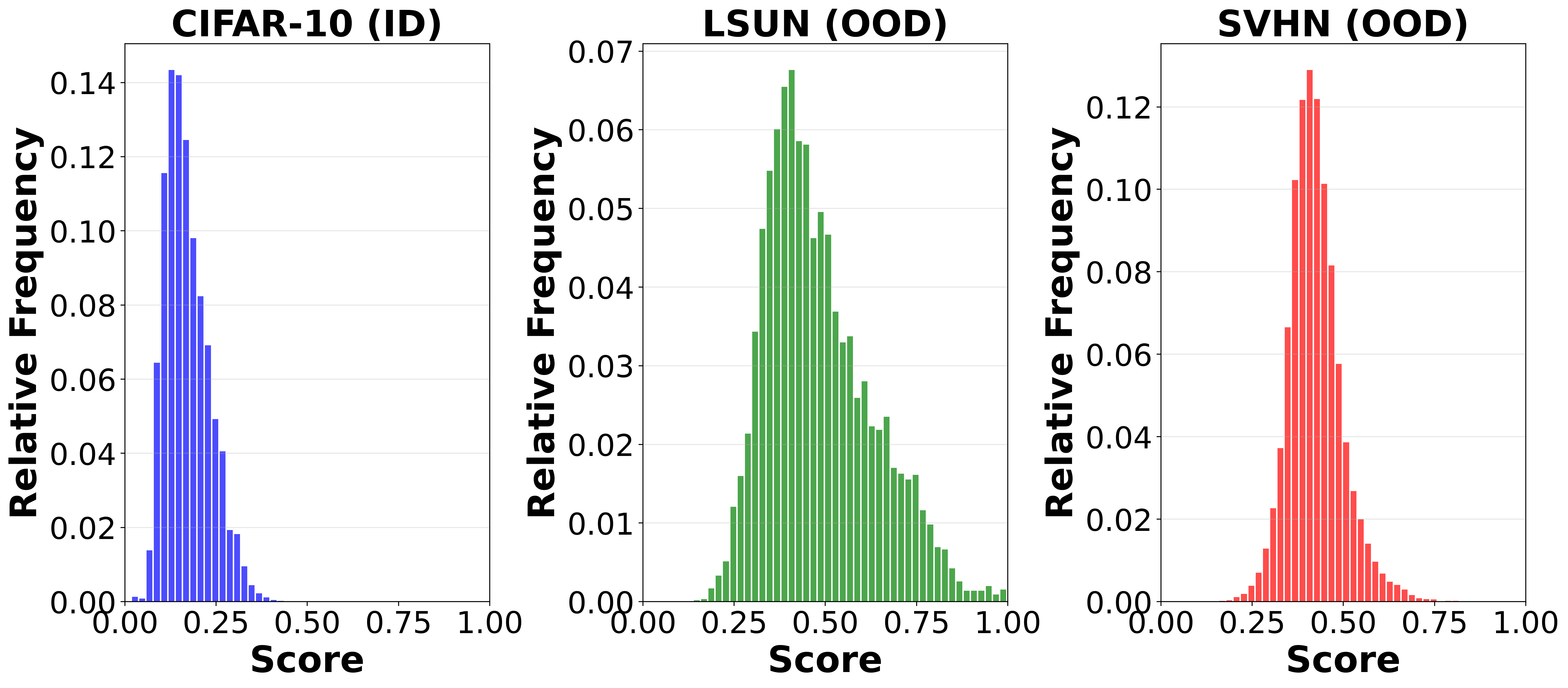}
        \caption{}
        \label{fig:histogram_knn_LSUN_SVHN}
    \end{subfigure}
    \hfill
    \begin{subfigure}[t]{0.48\linewidth}
        \centering
        \includegraphics[width=\linewidth]{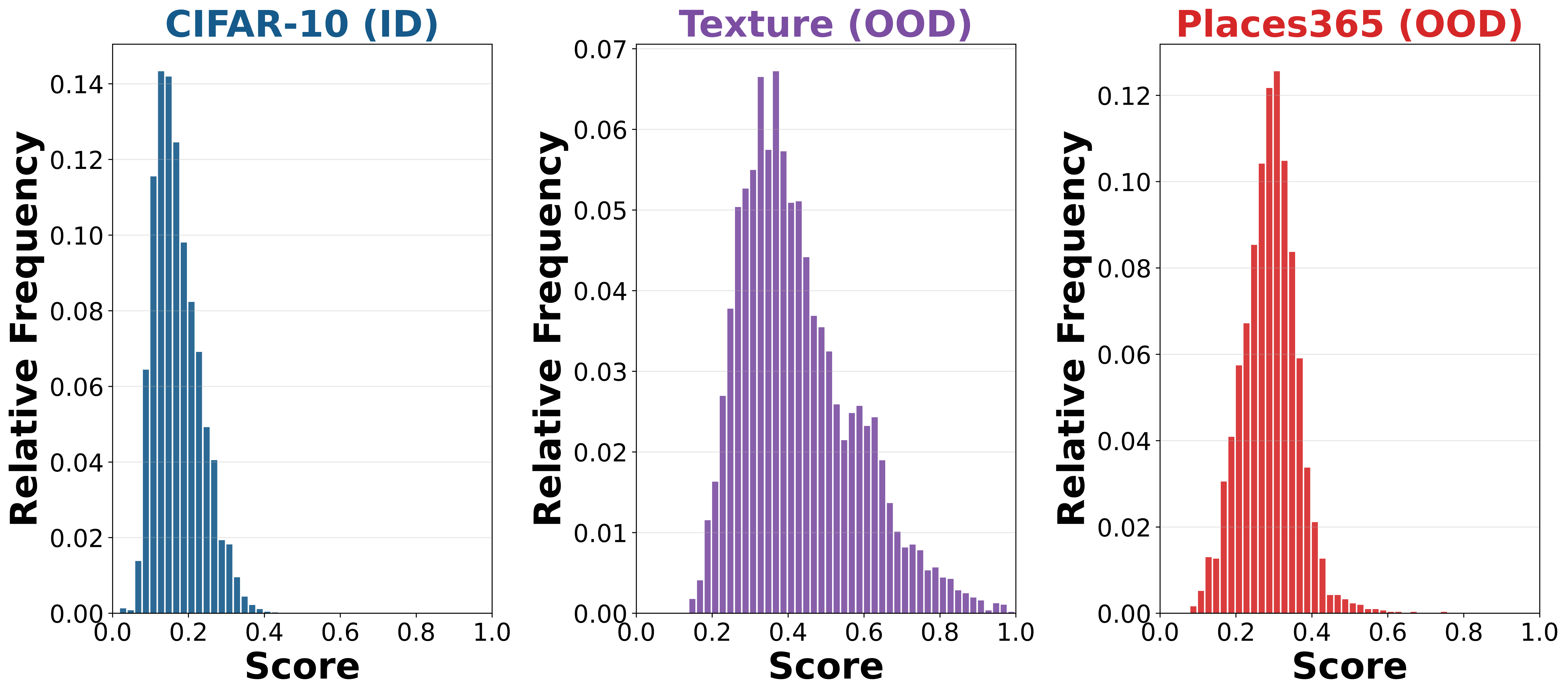}
        \caption{}
        \label{fig:histogram_knn_DTD_Places365}
    \end{subfigure}
    \caption{(a) $k$NN-based OOD detection score distribution for LSUN and SVHN. (b) $k$NN-based OOD detection score distribution for Texture and Places365. The ID dataset is CIFAR-10, and the encoder architecture is DenseNet-101.}
    \label{fig:knn-score-histogram}
\end{figure*}

\begin{figure*}[t]
    \centering
    \begin{subfigure}{0.47\linewidth}
        \centering
        \includegraphics[width=\linewidth]{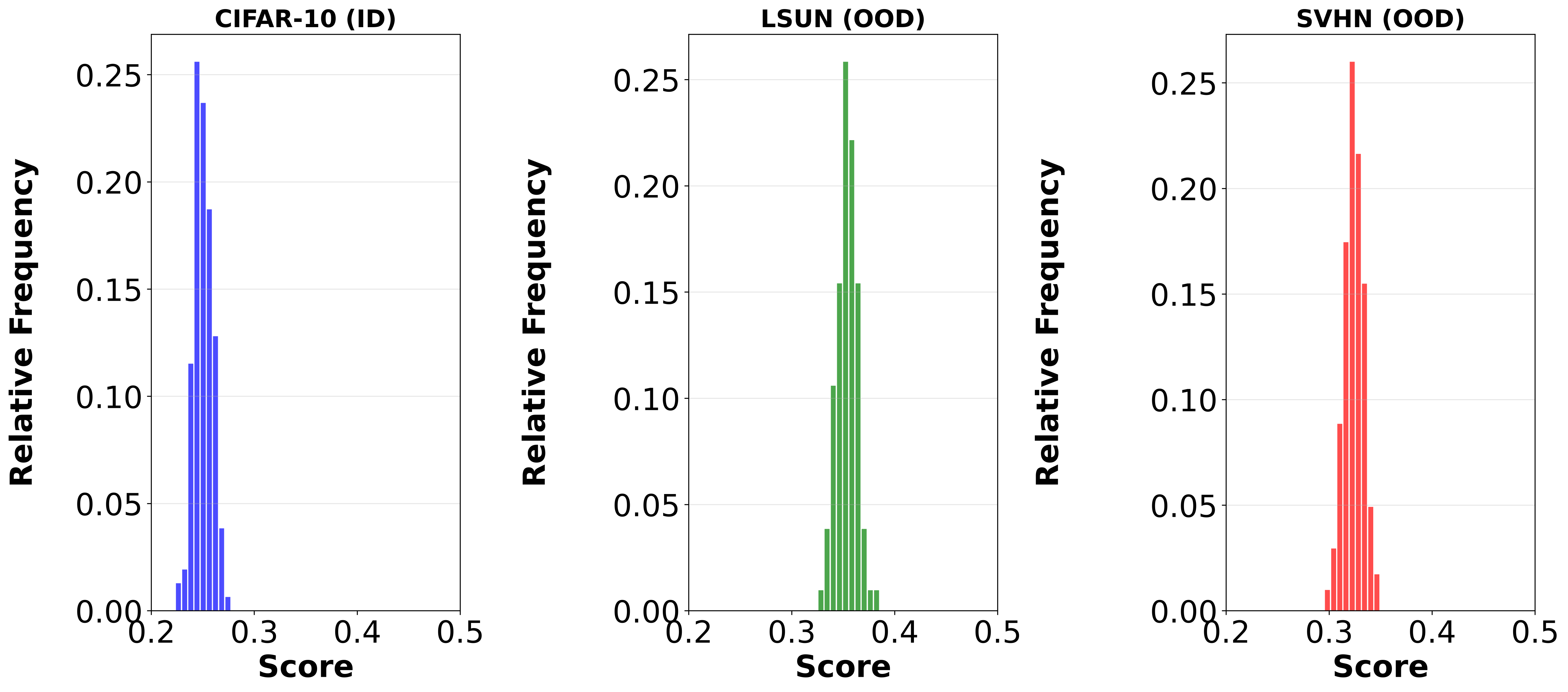}
        \caption{}
        \label{fig:histogram_reg}
    \end{subfigure}
    \hspace{0.04\linewidth} 
   \begin{subfigure}{0.47\linewidth} 
        \centering
        \includegraphics[width=\linewidth]{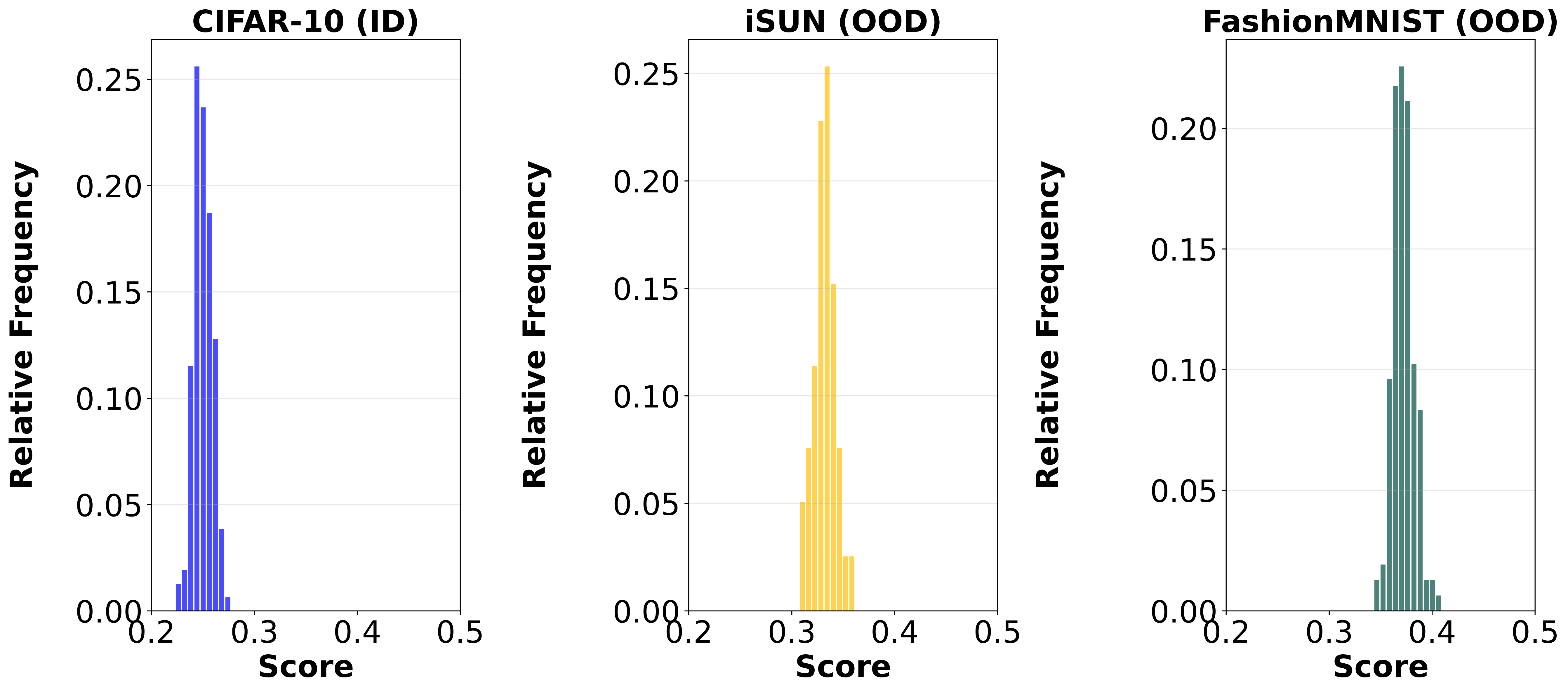}
        \caption{}
        \label{fig:histogram_reg_iSUN_FashionMNIST}
    \end{subfigure}

    \vspace{-0.3em} 
    
    \begin{subfigure}{0.47\linewidth} 
        \centering
        \includegraphics[width=\linewidth]{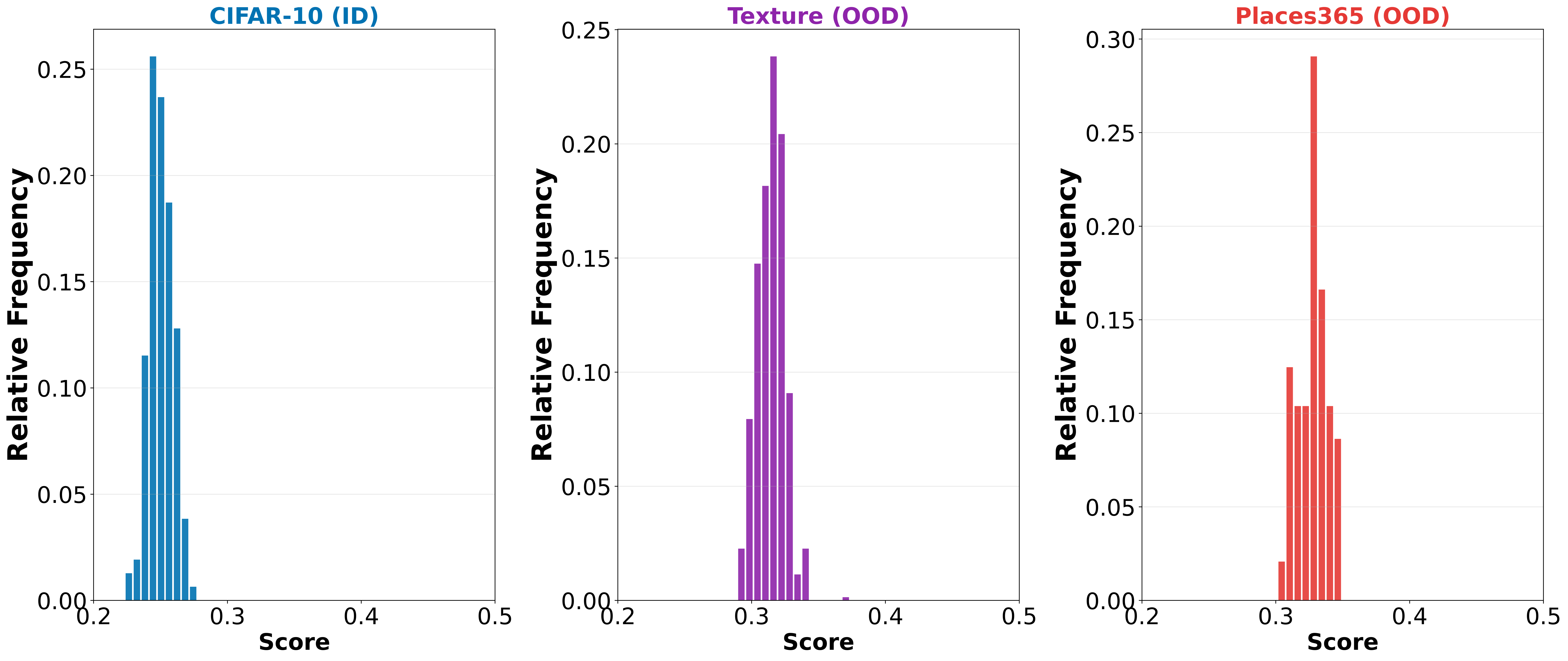}
        \caption{}
        \label{fig:histogram_reg_DTD_Places365}
    \end{subfigure}

    \caption{The subspace distance-based regularization loss (averaged over batch size) distribution: (a) SVHN and LSUN, (b) iSUN and FashionMNIST, (c) Texture and Places365. The ID dataset is CIFAR-10, and the encoder architecture is DenseNet-101.}
    \label{fig:reg-score-histogram}
\end{figure*}

\noindent
\textbf{Histogram Analysis.} In Fig. \ref{fig:knn-score-histogram}, we present the histogram of $k$NN-based detection scores for CIFAR-10, LSUN, SVHN, Texture, and Places365 datasets. Similar to iSUN and FashionMNIST, these plots exhibit a clear separation between ID and OOD features. Additionally, we present the histogram of the subspace distance-based regularization loss (averaged over batch) in Fig. \ref{fig:reg-score-histogram}. For every dataset, our regularization loss effectively provides a clear distinction between ID and OOD samples.

\begin{figure*}[htbp]
  \centering
  \captionsetup[subfigure]{justification=centering}

  \begin{subfigure}[t]{0.48\textwidth}
    \centering
    \includegraphics[height=6cm,keepaspectratio]{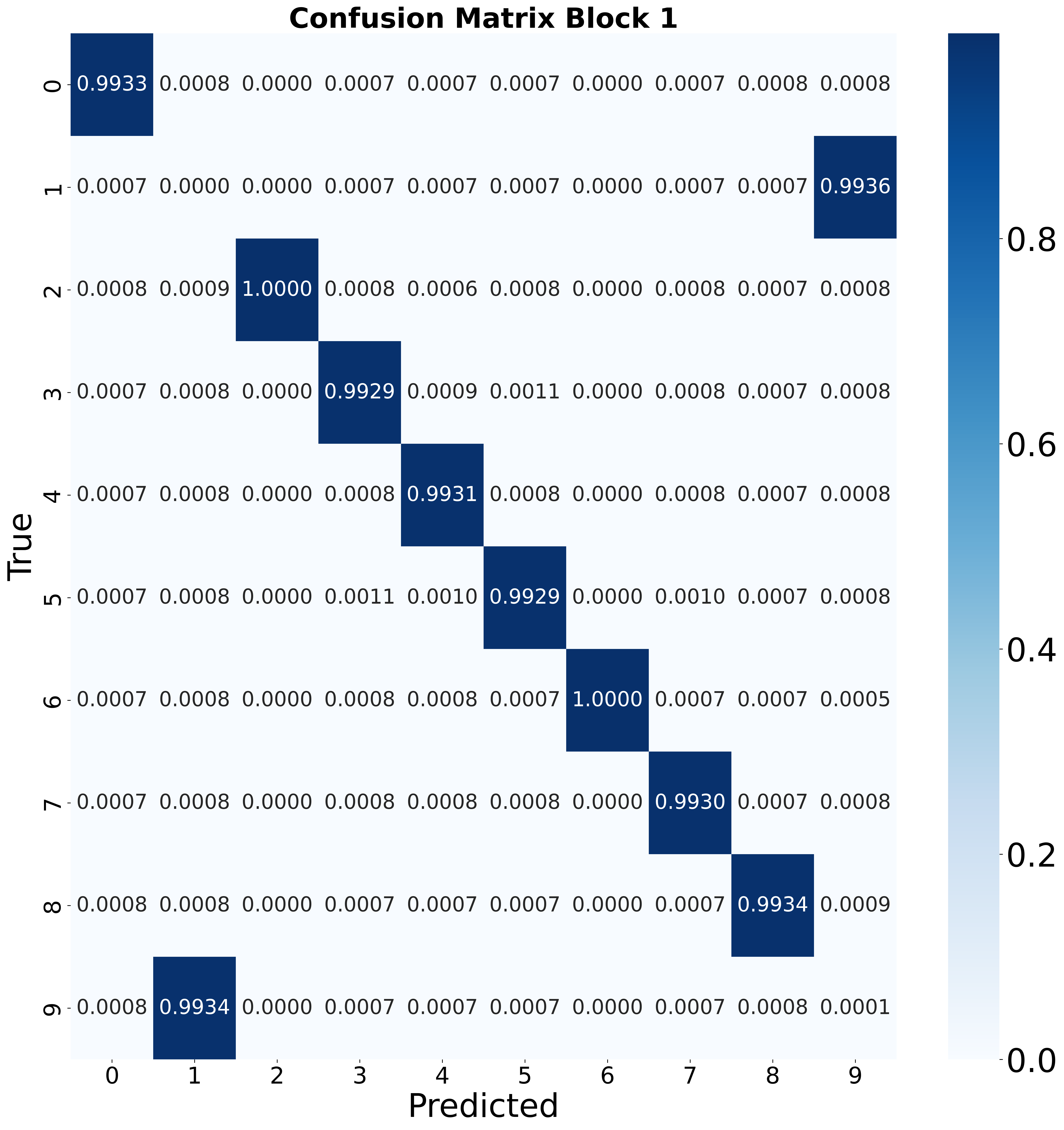}
    \caption{}
    \label{fig:cm1}
  \end{subfigure}
  \hfill
  \begin{subfigure}[t]{0.48\textwidth}
    \centering
    \includegraphics[height=6cm,keepaspectratio]{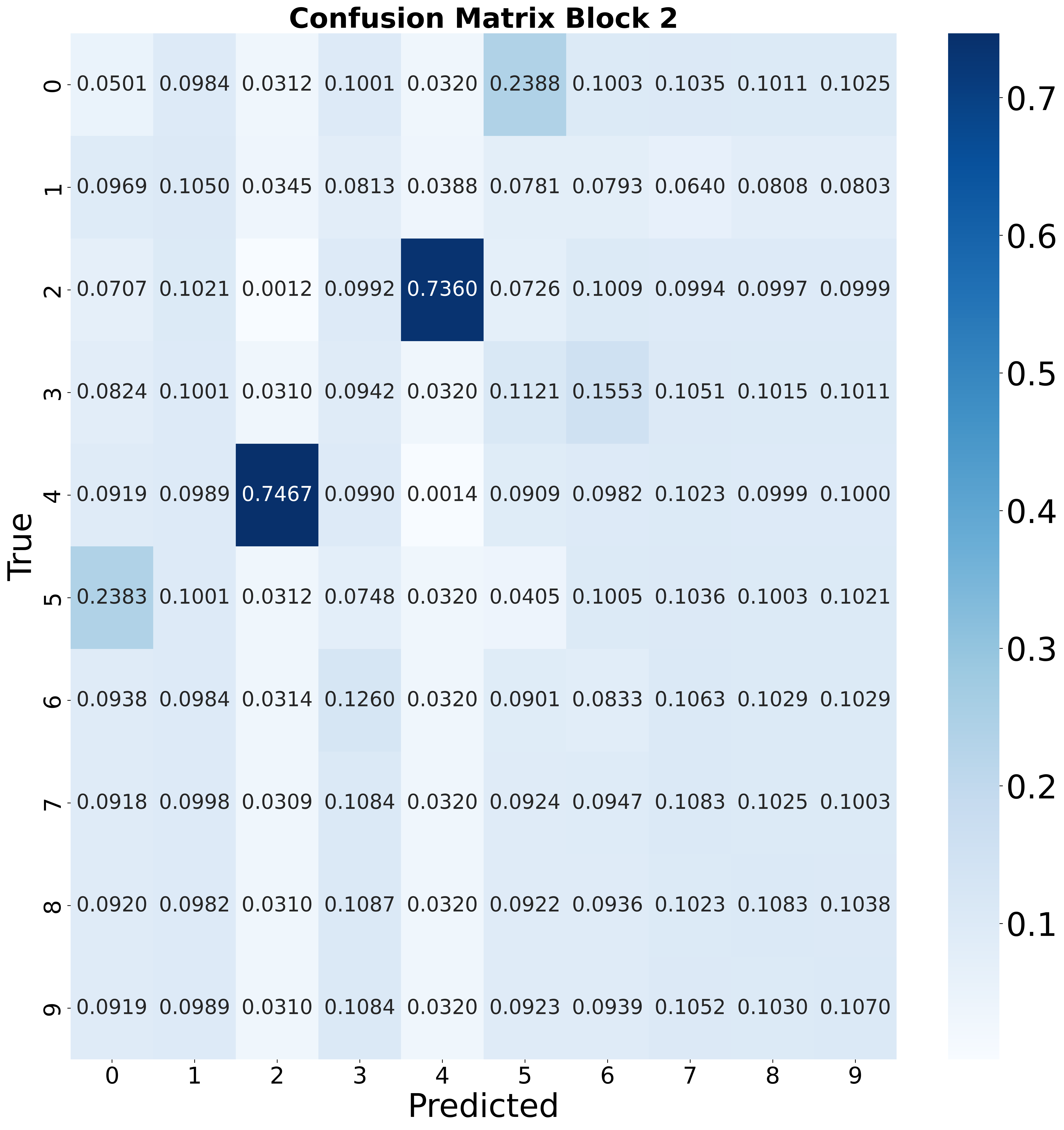}
    \caption{}
    \label{fig:cm2}
  \end{subfigure}
  \hfill
  \begin{subfigure}[t]{0.48\textwidth}
    \centering
    \includegraphics[height=6cm,keepaspectratio]{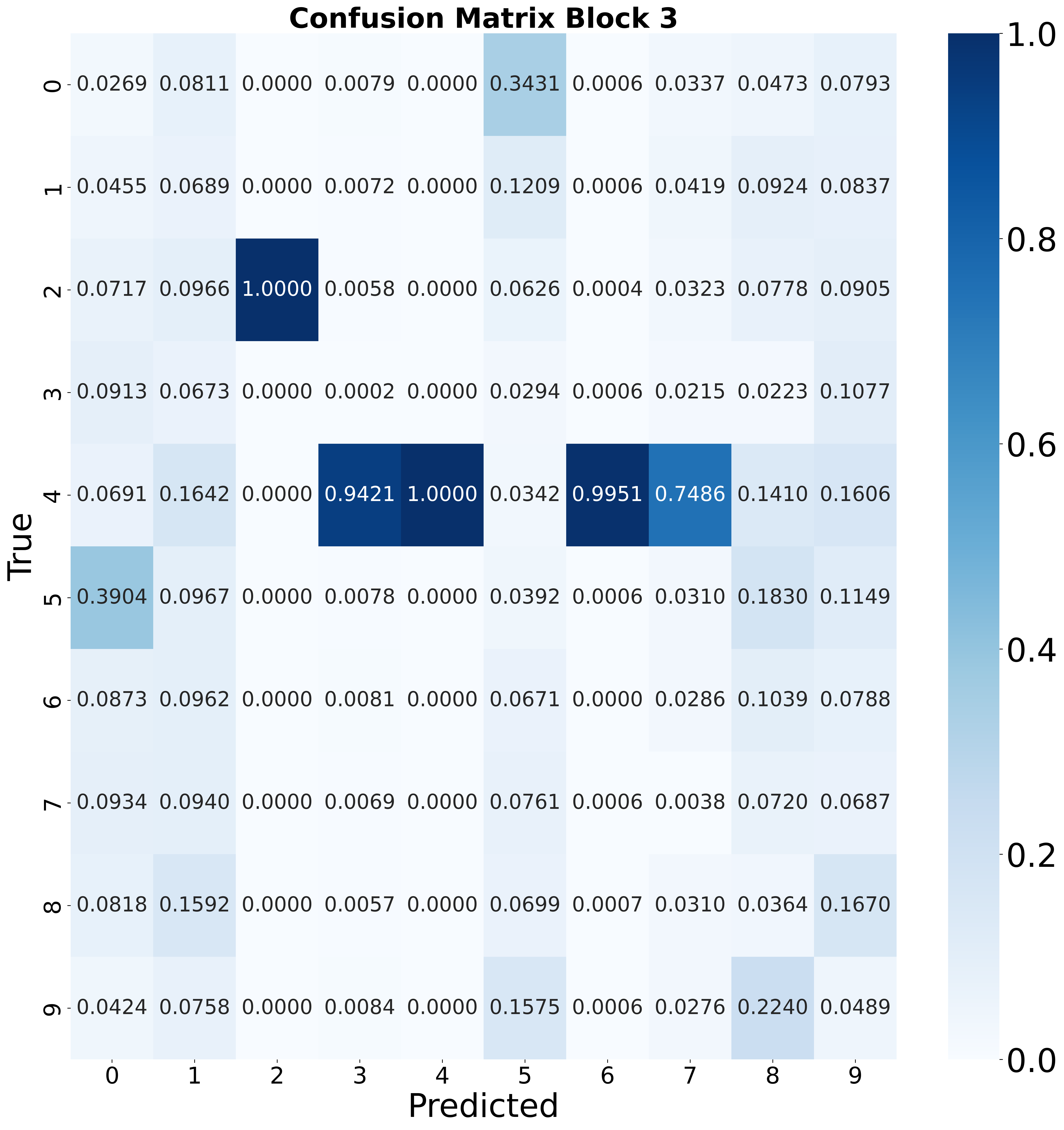}
    \caption{}
    \label{fig:cm3}
  \end{subfigure}

  \caption{Confusion matrices $\bm B_m$ learned by the PRISM approach for $M =6$.The ID dataset is CIFAR-10, and the encoder architecture is DenseNet-101.}
  \label{fig:confusion-matrices}
\end{figure*}

\noindent
\textbf{Confusion Matrices.} Fig.~\ref{fig:confusion-matrices} shows the confusion matrices learned by the PRISM approach. To generate these matrices, we set $M=6$ and $\lambda=0.1$ as hyperparameter settings. We used CIFAR-10 as the ID dataset and DenseNet-101 as the encoder architecture. One can note that the learned confusion matrices are distinct and do not lead to ``pseudo-label collapse"---allowing each pseudo label to contribute uniquely to define the subspace.

\newpage

\end{document}